\documentclass[runningheads]{llncs}
\usepackage[T1]{fontenc}
\usepackage{graphicx}
\usepackage{booktabs}
\usepackage[misc]{ifsym}
\newcommand{\corr}{(\Letter)}
\usepackage{mwe}

\usepackage{amsthm}
\usepackage{amsmath,amsfonts}
\usepackage{array}
\usepackage{subfig}
\usepackage{textcomp}
\usepackage{stfloats}
\usepackage{url}
\usepackage{verbatim}
\usepackage{graphicx}
\usepackage{balance}
\usepackage{hyperref}
\usepackage{eqnarray}
\usepackage{cite}
\usepackage{nicefrac}
\usepackage{bm}
\usepackage{amssymb}
\usepackage{mathtools}
\usepackage{bbm}
\usepackage{color}
\usepackage{caption}
\usepackage{algorithm}
\usepackage{appendix}
\usepackage[pdftex,dvipsnames]{xcolor}  
\usepackage[colorinlistoftodos,prependcaption]{todonotes}
\usepackage{algpseudocode}
\usepackage{dsfont}
\usepackage{stackengine}

\usepackage{tikz}
\usepackage[colorinlistoftodos]{todonotes}
\usepackage{xargs}                      
\usepackage[pdftex,dvipsnames]{xcolor}  

\newcommand{\E}{\mathbb{E}}

\newcommand{\R}{\mathbb{R}}

\newtheorem{assumption}{Assumption}

\begin{document}
\title{Collaborative Value Function Estimation Under Model Mismatch: A Federated Temporal Difference Analysis}

\titlerunning{Federated Temporal Difference Analysis Under Model Mismatch} 

\author{
Ali Beikmohammadi\inst{1} \corr \orcidID{0000-0003-4884-4600} 
\and 
Sarit Khirirat\inst{2} \orcidID{0000-0003-4473-2011} 
\and 
Peter Richt\'arik\inst{2} \orcidID{0000-0003-4380-5848} 
\and 
Sindri Magn\'usson\inst{1} \orcidID{0000-0002-6617-8683}
}


\authorrunning{A. Beikmohammadi et al.}
\institute{Department of Computer and Systems Sciences, \\
  Stockholm University, SE-164 25 Stockholm, Sweden \\
\email{\{beikmohammadi, sindri.magnusson\}@dsv.su.se}
\and
King Abdullah University of Science and Technology (KAUST), \\
Thuwal, Saudi Arabia \\
\email{\{sarit.khirirat, peter.richtarik\}@kaust.edu.sa}
}

\toctitle{Collaborative Value Function Estimation Under Model Mismatch: A Federated Temporal Difference Analysis} 
\tocauthor{Ali~Beikmohammadi,~Sarit~Khirirat,~Peter~Richt\'arik,~Sindri~Magn\'usson}

\maketitle             

\begin{abstract}
Federated reinforcement learning (FedRL) enables collaborative learning while preserving data privacy by preventing direct data exchange between agents. However, many existing FedRL algorithms assume that all agents operate in \emph{identical environments}, which is often unrealistic. In real-world applications, such as multi-robot teams, crowdsourced systems, and large-scale sensor networks, each agent may experience slightly different transition dynamics, leading to inherent model mismatches.  
In this paper, we first establish \emph{linear convergence} guarantees for single-agent temporal difference learning (\textsf{TD(0)}) in \emph{policy evaluation} and demonstrate that under a perturbed environment, the agent suffers a systematic bias that prevents accurate estimation of the true value function. This result holds under both \emph{i.i.d.} and \emph{Markovian} sampling regimes. We then extend our analysis to the federated TD(0) (\textsf{FedTD(0)}) setting, where multiple agents, each interacting with its own perturbed environment, \emph{periodically} share value estimates to collaboratively approximate the true value function of a common underlying model. Our theoretical results indicate the impact of model mismatch, network connectivity, and mixing behavior on the convergence of \textsf{FedTD(0)}.  
Empirical experiments corroborate our theoretical gains, highlighting that even moderate levels of information sharing significantly mitigate environment-specific errors.

\end{abstract}

\section{Introduction} \label{sec:Intro}
Reinforcement learning (RL) has been widely applied in various domains, including robotics, healthcare, finance, and game playing, where agents must learn to make sequential decisions in uncertain environments~\cite{SuttonBarto2018}. An agent in RL interacts with an unknown environment, observing states, taking actions, and receiving rewards, with the objective of learning an optimal policy that maximizes cumulative rewards. A key challenge in RL is learning the value function efficiently, especially when interactions with the environment are costly or time-consuming~\cite{Tesauro1995, beikmohammadi2023ta, beikmohammadi2024accelerating, beikmohammadi2023comparing}.

Recently, \emph{federated reinforcement learning} (FedRL) has emerged as a promising framework to improve sample efficiency and reduce learning time by allowing multiple agents to learn in parallel while exchanging information~\cite{Zhuo2019FRL, Qi2021FRLsurvey}. 
In a typical FedRL setup, multiple agents operate independently in separate instances of an environment, collect data locally, and communicate periodically their learned value estimates or policies to a central server or peer-to-peer network~\cite{Khodadadian2022FedRL, Blessing}. By aggregating these estimates, FedRL can improve learning efficiency and robustness, particularly in distributed applications that exhibit data privacy or communication constraints~\cite{Xie2021FedPG_BR, wu2021byzantine}.

However, existing literature in FedRL usually assumes that all agents interact with \emph{identical} environments \cite{doan2019finite, liu2023distributed, Khodadadian2022FedRL, beikmohammadi2024compressed, shen2023towards}. In practice, this assumption often fails to hold. In multi-robot teams, different robots may have slightly different actuators, sensors, or physical constraints, thus leading to variations in transition dynamics~\cite{Zhao2020SimToRealSurvey}. 
In crowdsourced RL tasks, different users experience distinct environments due to network conditions, device capabilities, or regional differences. Even in large-scale sensor networks, environmental fluctuations can introduce discrepancies in the transition dynamics observed by different sensors. These variations lead to \emph{model mismatch}, where each agent experiences a perturbed version of the \emph{true} environment~\cite{Pinto2017, Mankowitz2020}.

In this paper, we study whether agents operating under \emph{perturbed} transition dynamics can still collaboratively learn the true value function. Specifically, we focus on \emph{policy evaluation}, which plays a fundamental role in RL as a precursor to improve the policy~\cite{Tsitsiklis1997, Sutton1988, Bellman1957}. 
Our central research question is: \emph{Can agents collaboratively learn the true value function while each leveraging information from a potentially different noisy model?} The presence of model mismatch introduces systematic bias in the learned value function, which does not necessarily vanish with more iterations~\cite{Morimoto2005, jin2022federated, zhang2024finite, wang2023federated}. While policy evaluation is a natural starting point, recent works~\cite{srikant2019finite, even2003learning} have shown that with minor modifications, similar analysis techniques can extend to control settings.
\paragraph{Contributions.}
Our work provides a comprehensive theoretical analysis of temporal difference (TD) learning \cite{Sutton1988, SuttonBarto2018} under model mismatch in both \emph{single-agent} and \emph{federated} settings, under both \emph{i.i.d.} and \emph{Markovian} sampling regimes:
\begin{itemize}
    \item First, we analyze \textbf{single-agent \textsf{TD(0)} under  model mismatch}. Under both \emph{i.i.d.} and \emph{Markovian} sampling regimes, single-agent \textsf{TD(0)} attains linear convergence with the model mismatch error, which does not vanish as the algorithm progresses or when the step size decreases.
    \item We extend our analysis to \textbf{federated TD(0) (\textsf{FedTD(0)})}, where $N$ agents, each with a different transition kernel, periodically exchange value estimates. 
    \textsf{FedTD(0)} achieves linear convergence with the model mismatch error that is reduced by the aggregation of value functions evaluated by multiple agents. Furthermore, our convergence bounds explicitly depend on the number of agents, degree of heterogeneity, and communication frequency. 
    \item Finally, we corroborate our results with \textbf{numerical experiments}, showing that even moderate levels of collaboration significantly reduce bias and accelerate convergence to the true value function. 
    Our experiments’ code is publicly available at \url{https://github.com/AliBeikmohammadi/FedRL}.
\end{itemize}

\section{Related Work} \label{sec:Related}
\subsection{Single-agent TD Learning Algorithms} \label{sec:RelatedSingle-agentTD}
Much of the existing literature on TD learning has focused on the single-agent setting. 
Foundational studies proved the \emph{asymptotic} convergence of on-policy TD methods with function approximation, leveraging stochastic approximation theory \cite{tsitsiklis1996analysis, tadic2001convergence, borkar2009stochastic}.  
More recent advances have generalized these guarantees to off-policy learning scenarios \cite{maei2018convergent, zhang2020provably}.  
In parallel, a separate line of research has established \emph{finite-sample} or \emph{non-asymptotic} guarantees for TD learning.  
In the on-policy setting, TD learning  was studied under i.i.d. sampling regime by \cite{dalal2018finite, lakshminarayanan2018linear}, and Markovian sampling regime by \cite{bhandari2018finite, srikant2019finite, hu2019characterizing, chen2021lyapunov, mitra2024simple, samsonov2024improved}.
In the off-policy setting, non-asymptotic convergence of TD learning was also derived in \cite{chen2020finite, chen2021lyapunov}.  
Notably, most of these studies assume linear function approximation to leverage techniques from stochastic approximation.  
Our work departs from these single-agent analyses by examining a \emph{multi-agent federated} framework, where each agent interacts with its own perturbed environment and periodically shares updates with others.  This setting poses additional challenges related to agent heterogeneity, multiple dynamics, and intermittent communication, thereby requiring a specialized analytical approach.

\subsection{Distributed and Federated RL Algorithms}\label{sec:RelatedFederatedRL}
\paragraph{Distributed RL.}
Many distributed RL algorithms have emerged to address scalability and efficiency challenges. 
This progress is exemplified by the development of many distributed frameworks, such as asynchronous parallelization \cite{mnih2016asynchronous}, the high-throughput IMPALA architecture \cite{espeholt2018impala}, and decentralized gossip-based protocols \cite{assran2019gossip}.
Theoretical convergence properties of such distributed methods have expanded to include robustness against adversarial attacks \cite{wu2021byzantine, Xie2021FedPG_BR} and communication compression \cite{mitra2023temporal}. 
Furthermore, several works derived sample complexities under i.i.d. sampling for distributed RL with linear function approximation \cite{doan2019finite, liu2023distributed}, and actor-critic algorithms \cite{chen2022sample, shen2023towards}.
Further work extends to decentralized stochastic approximation \cite{sun2020finite, wai2020convergence, zheng2023federated}, TD learning with linear function approximation \cite{wang2023federated}, and off-policy TD actor-critic algorithms \cite{chen2021multi}.
However, these analyses commonly rely on two key assumptions: agents synchronize updates through continuous communication (e.g., after every local iteration) and operate in \emph{homogeneous} environments with identical dynamical properties across all participants.

\paragraph{FedRL under Homogeneous Environments.}
Parallel efforts in FedRL aim to reduce communication costs by allowing agents to perform multiple local updates between periodic synchronization rounds. This paradigm has been explored in contexts such as federated TD learning with linear function approximation \cite{dal2023federated, dal2023over, Khodadadian2022FedRL}, off-policy TD methods \cite{Khodadadian2022FedRL}, and $Q$-learning \cite{Khodadadian2022FedRL}, with further extensions to behavior-policy heterogeneity \cite{woo2023blessing} and compressed communication \cite{beikmohammadi2024compressed}. However, a common limitation across all these approaches remains persistent: all agents are assumed to operate in identical Markov decision processes (MDPs) (\emph{homogeneous} environments). This assumption fails to capture real-world scenarios where agents face environmental heterogeneity.

\paragraph{FedRL under Heterogeneous Environments.}
To our knowledge, only four works have studied FedRL under \emph{heterogeneous} environments—a more closely related to our setting \cite{jin2022federated, xie2023fedkl, zhang2024finite, wang2023federated}, but each differs substantially from ours.  
For instance, \cite{jin2022federated} focuses on tabular $Q$-learning, and \cite{xie2023fedkl} studies the asymptotic behavior of distributed $Q$-learning.
Meanwhile, \cite{zhang2024finite} considers SARSA with linear function approximation, modeling data heterogeneity via a worst-case total-variation distance between transition kernels.
The most closely related work is \cite{wang2023federated}, which analyzes TD learning under Markovian sampling and periodic communication in a heterogeneous setting, but the analysis relies on function approximation, employs a more restrictive multiplicative bound on transition-model deviation, and uses a norm-induced approach that differs from ours. 
We depart from these approaches by considering a tabular federated policy evaluation problem in which each agent's environment is drawn from a distribution centered on the true transition model, and the agents periodically communicate with a central server.  Our framework handles both i.i.d. and Markovian sampling, and it explicitly quantifies how inter-agent collaboration mitigates the bias introduced by environment perturbations in estimating the true value function.  To our knowledge, this is the first systematic treatment of \emph{model mismatch} in a tabular RL setting for both the single-agent and federated scenarios.

\section{Model and Background} \label{sec:Background}
Within this section, we formalize the MDP setting and key definitions. 

\paragraph{Discounted Infinite-horizon MDP.}
We study a discounted infinite-horizon MDP formally defined as the tuple $\mathcal{M} \;=\; \langle \mathcal{S}, \mathcal{A}, \mathcal{R}, \mathcal{P}, \gamma\rangle$. Here, \(\mathcal{S}\) and  \(\mathcal{A}\) are finite state and action spaces, respectively; \(\mathcal{P}\) represents a set of the action-dependent transition probabilities; \(\mathcal{R}\) is a bounded reward function; \(\gamma \in (0,1)\) is the discount factor governing long-term reward trade-offs.

\paragraph{Value Function and Bellman Operator.}
Under a fixed policy \(\mu\colon \mathcal{S}\to \mathcal{A}\), the MDP reduces to a Markov reward process (MRP) with transition matrix \(P_\mu\) and reward function \(r := R_\mu\). Here, \(P_\mu(s, s')\) is the probability of transitioning from \(s\) to \(s'\) under action \(\mu(s)\), and \(r(s)\) denotes the expected immediate reward at state \(s\). Our goal is to evaluate the value function \(V\), which captures the expected discounted return when starting from any state \(s\) and following~\(\mu\):
\begin{equation}
V(s)
\;=\;
\mathbb{E}\Bigl[\,
\sum_{t=0}^{\infty} \gamma^t\,r(s^{(t)})
\;\big|\;
s^{(0)} = s
\Bigr],
\quad
\forall s \in \mathcal{S},
\end{equation}
where the expectation is taken over trajectories generated by the transition kernel $P_\mu$ (i.e., $s^{(t+1)} \sim P_\mu\left(\cdot \mid s^{(t)}\right)$ ) for all $t \geq 0$. 

A fundamental result in dynamic programming \cite{Bellman1957, bertsekas1996neuro} states that \(V\) is the unique fixed point of the policy-specific Bellman operator \(T \colon \mathbb{R}^{|\mathcal{S}|} \to \mathbb{R}^{|\mathcal{S}|}\), defined as:
\begin{equation}
\bigl(TV\bigr)(s)
\;=\;
r(s)
\;+\;
\gamma\,\sum_{s' \in \mathcal{S}} P_\mu(s, s')\,V(s')
\quad
\forall s \in \mathcal{S}.
\end{equation}
Equivalently, \(V\) satisfies the Bellman equation:
$TV = V$.

\section{Single-Agent \textsf{TD(0)} under Model Mismatch} \label{sec:SATD}
One popular approach for estimating the value function $V$ in a model-free setting (i.e., when the underlying MDP is unknown) is the \textsf{TD(0)} algorithm \cite{Sutton1988, SuttonBarto2018}.  
At each time step $t$, the agent observes a state $s^{(t)}$, receives reward $r^{(t)} := r(s^{(t)})$, transitions to a next state $s^{(t+1)}$, and updates its current value estimate $V^{(t)}(\cdot)$ by modifying only the coordinate at $s^{(t)}$. 
Specifically, \textsf{TD(0)} updates via: 
\begin{equation}\label{eq:TD0-update}
V^{(t+1)}\bigl(s^{(t)}\bigr) 
\;=\;
V^{(t)}\bigl(s^{(t)}\bigr) 
\;+\; 
\alpha \,\delta^{(t)} \text{~~and~~} \delta^{(t)}=
    r^{(t)} 
    \;+\;
    \gamma\,V^{(t)}\bigl(s^{(t+1)}\bigr)
    \;-\;
    V^{(t)}\bigl(s^{(t)}\bigr),
\end{equation}
where $\alpha \in (0,1)$ is a step size, and $V^{(0)}$ is some initial guess. For states $s \neq s^{(t)}$, we simply set $V^{(t+1)}(s) = V^{(t)}(s)$. Under standard assumptions—such as a sufficiently small step size \(\alpha\), bounded rewards, and ergodic state sampling—\textsf{TD(0)} converges to the unique fixed point of the Bellman operator \(T\), which is precisely the true value function \(V\) of the policy \(\mu\) \cite{tsitsiklis1996analysis, borkar2009stochastic, SuttonBarto2018}.  

However, this holds when the agent interacts with the true transition probability \( P_\mu \), which we aim to evaluate the policy for.
Here, we instead consider scenarios where the agent interacts with a perturbed environment \(\hat P\) rather than \(P_\mu\), as explained in Algorithm~\ref{alg:single_agent_TDlearning}.
Specifically, we analyze the convergence of the single-agent \textsf{TD(0)} algorithm under the following model mismatch assumption.
\begin{assumption}\label{assum:hat_P_single} Let  the empirical transition matrix $\hat P$ be a perturbed transition matrix of $P_\mu$, which satisfies for $\Delta >0$
	\begin{eqnarray}\label{eqn:hat_P_single}
		\left\| \hat P - P_\mu \right\|^2_{2} \leq \Delta^2.
	\end{eqnarray}
\end{assumption}	
Assumption~\ref{assum:hat_P_single} implies the deviation of the empirical transition matrix $\hat P$ from the true transition matrix $P_\mu$. Small values of $\Delta$ imply that $\hat P$ is close to $P_\mu$. 
Furthermore, $\hat P$ can possibly be a biased estimate of $P_\mu$.

\begin{algorithm}[t] 
\caption{Single-agent \textsf{TD(0)}}\label{alg:single_agent_TDlearning}
\begin{algorithmic}
\State \textbf{Initialize:} Learning rate $\alpha \in (0,1)$, Discount factor $\gamma \in (0,1)$, Number of communication rounds $T$, Initial value function $V^{(0)}$.
\For{each time step $t=0,1,\ldots,T-1$} 
    \State Observe state $s^{(t)}$ according to the chosen sampling regime (i.i.d. or Markovian).  
    \State Receive $r^{(t)}$ and get $s^{(t+1)} \sim \hat P (\cdot | s^{(t)})$.
    \State Update $V^{(t+1)}$ via $$V^{(t+1)}(s^{(t)})= V^{(t)}(s^{(t)}) + \alpha [r^{(t)} + \gamma V^{(t)}(s^{(t+1)})-V^{(t)}(s^{(t)})].$$
\EndFor   
\end{algorithmic}
\end{algorithm}

Next, we analyze the convergence of the single-agent \textsf{TD(0)} algorithm under model mismatch with both well-known sampling regimes, i.i.d. and Markovian, in Sections \ref{sec:SA_IID} and \ref{sec:SA_Markov}, respectively.  

\subsection{Analysis for I.I.D. Setting} \label{sec:SA_IID}
In the i.i.d.\ sampling regime, each state \( s^{(t)} \) is independently drawn from the stationary distribution over \( \mathcal{S} \).  
Although \( s^{(t)} \) and \( s^{(t-1)} \) are uncorrelated, we still generate a next state \( s^{(t+1)} \sim \hat P(\cdot \mid s^{(t)}) \) in \textsf{TD(0)} update~\eqref{eq:TD0-update} to perform one-step bootstrapping.  
Hence, while the \emph{tuples} \( (s^{(t)}, s^{(t+1)}) \) are sampled i.i.d. from the stationary distribution, the actual \emph{visited} states do not form a Markov chain.  
The i.i.d. sampling regime eliminates temporal correlations in the sampled states, and therefore simplifies the convergence analysis.

The following theorem establishes the linear convergence of single-agent \textsf{TD(0)} under i.i.d. sampling.

\begin{theorem}[Single-agent, i.i.d. sampling]\label{thm:single_iid}
Consider the single-agent \textsf{TD(0)} algorithm (Algorithm~\ref{alg:single_agent_TDlearning}) under the i.i.d. sampling. 
Let a row-stochastic transition matrix $\hat P$ satisfy Assumption~\ref{assum:hat_P_single}. 
Then, with probability at least $1-\delta$,
\begin{eqnarray*}
	\| e^{(t)} \|_2 
	& \leq &  [1-\alpha(1-\gamma)]^t \|  e^{(0)} \|_2 +  \frac{\gamma\Delta \sqrt{|\mathcal{S}|} }{(1-\gamma)^2}  + \frac{\alpha \sqrt{t}}{(1-\gamma)}\sqrt{32(\log(t/\delta)+1/4)}, 
\end{eqnarray*}
where $e^{(t)}:= V^{(t)}-V$, $|\mathcal{S}|$ is the number of states, and $\alpha\in(0,1)$.
\end{theorem}

\begin{remark}
From Theorem~\ref{thm:single_iid}, the single-agent \textsf{TD(0)} algorithm ensures linear convergence with a high probability of its value function $V^{(t)}$ towards the unique true value function $V$ with the first error term due to the model mismatch $\Delta$ and the second error term of $\frac{\alpha \sqrt{t}}{(1-\gamma)}\sqrt{32(\log(t/\delta)+1/4)}$ due to the i.i.d. sampling regime.
Decreasing the step size $\alpha$ cannot reduce the first error term due to the model mismatch, while decreasing the second error term due to i.i.d. sampling at the price of a slow convergence rate.
For instance, if we choose  $\alpha=1/T^\eta$ with $\eta \in (1/2,1)$, $0<t\leq T$, where $T$ is the given number of total iteration counts, then the algorithm converges at the rate:
\begin{eqnarray*}
    \| e^{(t)} \|_2 
    & \leq & \exp\left(-\frac{t(1-\gamma)}{T^\eta} \right)\|  e^{(0)} \|_2 + \mathcal{O}\left( \frac{\sqrt{t \log(t)}}{T^{\eta}} \right) + { \frac{\gamma\Delta \sqrt{|\mathcal{S}|}}{(1-\gamma)^2}},
\end{eqnarray*}
with probability at least $1-\delta$. In conclusion, the convergence bound for the algorithm contains the model mismatch error term ${ \frac{\gamma\Delta \sqrt{|\mathcal{S}|}}{(1-\gamma)^2}}$.
\end{remark}

\subsection{Analysis for Markovian Setting} \label{sec:SA_Markov}
As a more realistic setting, we also study Algorithm~\ref{alg:single_agent_TDlearning} under a Markovian sampling regime.  
In this regime, the states follow a Markov chain defined by the transition matrix $\hat P$. Concretely, $s^{(t+1)} \sim \hat P(\cdot \mid s^{(t)})$ for each $t$, matching our earlier definition of MRP induced by a fixed policy. Although this better captures real-world dynamics, analyzing the corresponding \textsf{TD(0)} updates becomes more intricate due to the temporal correlations in $\{s^{(t)}\}$.
To analyze the single-agent \textsf{TD(0)} algorithm under the Markovian sampling regime, we make an assumption on the Markov chain and define the mixing time $\tau_{\epsilon}$ in the single-agent setting.
\begin{assumption}\label{assumption:aperiodic_irreducible}
The Markov chain induced by the policy $\mu$ is aperiodic and irreducible.     
\end{assumption}
\begin{definition}\label{def:mixing_time}
Let $\tau_{\epsilon}$ be the minimum time such that the following holds: 
\(
    \| \xi^{(t)} \|_2 \leq \epsilon,  \forall t \geq \tau_{\epsilon}. 
\)
Here, $\xi^{(t)}\in\R^{|\mathcal{S}|}$ is defined by
$\xi^{(t)}(s)=\delta^{(t)}-\mathbb{E}\bigl[\delta^{(t)} \mid \mathcal{F}^{(t)}\bigr]$ when $s=s^{(t)}$, and $\xi^{(t)}(s)=0$ for all other $s$, where $\mathcal{F}^{(t)}$ is the filtration up to $t$.
\end{definition}
This assumption implies that the Markov chain induced by $\mu$ admits a unique stationary distribution $\pi$, and mixes at a geometric rate~\cite{levin2017markov}. The consequence of this assumption is that there exists some $T \geq 1$ such that $\tau_{\epsilon} \leq T \log(1/\epsilon)$.  

The next theorem shows the linear convergence of the single-agent \textsf{TD(0)} algorithm under Markovian sampling. 

\begin{theorem}[Single-agent, Markovian sampling] \label{thm:single_markov}
Consider the single-agent \textsf{TD(0)} algorithm (Algorithm~\ref{alg:single_agent_TDlearning}) under Markovian sampling. 
Let a row-stochastic transition matrix $\hat P$ satisfy Assumption~\ref{assum:hat_P_single} and the Markov chain satisfy Assumption~\ref{assumption:aperiodic_irreducible}.
Then, 
\begin{eqnarray*}
	\| e^{(t)} \|_2 
	\leq [1-\alpha(1-\gamma)]^t \|  e^{(0)} \|_2 + \gamma \frac{\Delta \sqrt{|\mathcal{S}|} }{(1-\gamma)^2} + \frac{\alpha}{1-\gamma} \left( 2\tau_{\alpha}+1\right),
\end{eqnarray*}
where $e^{(t)}:= V^{(t)}-V$, $|\mathcal{S}|$ is the number of states, $\alpha\in(0,1)$, and $\tau_\alpha$ is defined by Definition~\ref{def:mixing_time}.
\end{theorem}

\begin{remark}
The single-agent \textsf{TD(0)} algorithm under the Markovian sampling regime, similar to the i.i.d. sampling regime,  achieves the linear convergence of its value function $V^{(t)}$ towards the fixed value function $V$ with two residual error terms. 
In particular, decreasing the step size $\alpha$ cannot reduce the first error term due to the model mismatch $\Delta$, while decreasing the second error term due to the mixing time $\tau_\alpha$ of the aperiodic and irreducible Markov chain at the price of slow convergence. 
For instance, the algorithm with $\alpha = 1/T^{\eta}$ for $\eta\in(1/2,1)$  converges at the rate
\begin{eqnarray*}
    \| e^{(t)} \|_2 
    & \leq & \exp\left(-\frac{t(1-\gamma)}{T^\eta} \right)\|  e^{(0)} \|_2 + \mathcal{O}\left( \frac{1}{T^{\eta}} \right) + { \frac{\gamma\Delta \sqrt{|\mathcal{S}|} }{(1-\gamma)^2}}.
\end{eqnarray*}
In conclusion, the single-agent \textsf{TD(0)} algorithm under both i.i.d sampling and Markovian sampling regimes converges with the model mismatch error term $ { \frac{\gamma\Delta \sqrt{|\mathcal{S}|} }{(1-\gamma)^2}}$ that cannot be reduced by decreasing the step size.
In the next section, we show that this model mismatch error term can be reduced by the benefits of multiple agents for running the \textsf{TD(0)} algorithm in the federated setting. 
\end{remark}

\section{Extension to \textsf{FedTD(0)} under Model Mismatch} \label{sec:MATD}
To show the benefits of multiple agents, we extend our results to the federated setting. 
We consider the \textsf{FedTD(0)} algorithm, where multiple agents collaboratively estimate the value function. 
In each communication round $t=0,1,\ldots,T-1$, each agent $i = 1,2,\ldots,N$ receives the global value function estimate $V^{(t)}$ from the server, sets $V_i^{(t,0)}=V^{(t)}$, and updates its local estimate $V_i^{(t,k)}$ according to: 
\begin{eqnarray*}
V_i^{(t,k+1)}(s_i^{(t,k)}) &=& V_i^{(t,k)}(s_i^{(t,k)}) + \alpha \delta_i^{(t,k)},  \quad \text{and} \\
\delta_i^{(t,k)} &=& r_i^{(t,k)} + \gamma V_i^{(t,k)}(s_i^{(t,k+1)})-V_i^{(t,k)}(s_i^{(t,k)}) ~ \text{for} ~ k=0,1,{\ldots},K{-}1.
\end{eqnarray*}
Here, the step size is denoted by \( \alpha \in (0,1) \), \( s_i^{(t,k)} \) is the observed state, \( r_i^{(t,k)} \) is the received reward for each agent, and \( s_i^{(t,k+1)} \) is drawn from the agent's transition probability conditioned on \( s_i^{(t,k)} \).
Then, the central server computes the average of the received estimate progress from all the agents $\frac{1}{N}\sum_{i=1}^N (V_i^{(t,K)}-V^{(t)})$, and updates the global value function estimate via:
\begin{eqnarray*}
    V^{(t+1)} = V^{(t)} + \frac{\beta}{N}\sum_{i=1}^N (V_i^{(t,K)}-V^{(t)}),
\end{eqnarray*}
where $\beta \in (0,1]$ is the federated tuning parameter. 
The description of \textsf{FedTD(0)} algorithm is provided in Algorithm~\ref{alg:fed_TDlearning}.
To demonstrate the benefit of multiple agents in \textsf{FedTD(0)}, we impose the following model mismatch assumption.
\begin{assumption}\label{assum:hat_P}
Let the empirical transition matrices $\hat P_1,\hat P_2,\ldots,\hat P_N$ be the perturbed matrix of  $P_\mu$, which satisfies Assumption~\ref{assum:hat_P_single} with $\Delta>0$,
 and also
\begin{eqnarray}\label{eqn:hat_P}
    \left\| \frac{1}{N}\sum_{i=1}^N \hat P_i - P_\mu \right\|^2_{2} \leq \frac{\Lambda^2}{N}.
\end{eqnarray}
\end{assumption}
This assumption implies that as the number of agents $N$ grows, the average model $(1/N)\sum_{i=1}^N \hat{P}_i$ converges to the true model $P_{\mu}$ at the rate $\mathcal{O}(1/N)$. This captures the intuition that if agents have similar dynamics on average, their collective behavior also becomes more predictable.  
In particular, this assumption captures well when $\hat P_i$ are row-stochastic, i.e. $\| \hat P_i \|_2 \leq \| \hat P_i \|_1 \leq 1$, and are sampled under both i.i.d. and Markovian sampling regimes. 
On the one hand, if $\hat P_i$ are sampled under i.i.d. sampling, and satisfy $\E\|\hat{P}_i\|_2=P_{\mu}$ and $\| \hat{P}_i \|_2\leq 1$, then according to Lemma A.2. of \cite{dorfman2022adapting}, we obtain \eqref{eqn:hat_P} with $\Lambda = [6(1+\sqrt{\log(1/\delta)})]^2$ with probability at least $1-\delta$.
On the other hand, if $\hat P_i$ are sampled under Markovian sampling and satisfy $\| \hat P_i \|_2 \leq 1$, and the Markov chain satisfies Assumption~\ref{assumption:aperiodic_irreducible} with the mixing time $\tau_{\epsilon}$, then according to Lemma A.6 of \cite{dorfman2022adapting} we have \eqref{eqn:hat_P} with $\Lambda = \mathcal{\tilde O}\left(   \tau_{\epsilon}\lceil 2\log(N)\rceil \right)$ in expectation.
\begin{algorithm}[t]
\caption{\textsf{FedTD(0)}}\label{alg:fed_TDlearning}
\begin{algorithmic}
\State \textbf{Initialize:} Learning rate $\alpha \in (0,1)$, Federated parameter $\beta \in (0,1]$, Discount factor $\gamma \in (0,1)$, Number of agents $N$, Number of local steps $K$, Number of communication rounds $T$, Initial value function $V^{(0)}$.
\For{each communication round $t=0,1,\ldots,T-1$} 
   \For{each agent $i=1,2,\ldots,N$ \textbf{in parallel}} 
        \State Set $V_i^{(t,0)}=V^{(t)}$, where $V^{(t)}$ is the global value estimate from the server.  
        \For{$k=0,1,\ldots,K-1$}
        \State Observe state $s_i^{(t,k)}$ according to the chosen sampling  (i.i.d. or Markovian). 
        \State Receive $r_i^{(t,k)}$ and get $s_i^{(t,k+1)} \sim \hat P_i (\cdot | s_i^{(t,k)})$.
        \State Update $V_i^{(t,k+1)}$ via $$V_i^{(t,k+1)}(s_i^{(t,k)})= V_i^{(t,k)}(s_i^{(t,k)}) + \alpha[r_i^{(t,k)} + \gamma V_i^{(t,k)}(s_i^{(t,k+1)})-V_i^{(t,k)}(s_i^{(t,k)})].$$
        \EndFor
        \State Send $V_i^{(t,k+1)} - V^{(t)}$ back to the server.
   \EndFor
   \State Server computes and broadcasts global $$ V^{(t+1)}  = V^{(t)} + \frac{\beta}{N}\sum_{i=1}^N (V_i^{(t,K)}-V^{(t)}).$$
\EndFor   
\end{algorithmic}
\end{algorithm}

\subsection{Analysis for I.I.D. Setting} \label{sec:MA_IID}
\textsf{FedTD(0)} under i.i.d. sampling achieves linear convergence in high probability with a lower residual error than single-agent \textsf{TD(0)}, as shown next. 

\begin{theorem}[Federated, i.i.d. sampling] \label{thm:multi_iid}
Consider \textsf{FedTD(0)} (Algorithm~\ref{alg:fed_TDlearning}) under i.i.d. sampling. Let each row-stochastic transition matrix $\hat P_i$ of the $i^{\rm th}$-agent  satisfy Assumption~\ref{assum:hat_P}. Then, 
with probability at least $1-\delta$, 
\begin{eqnarray*}
    &&	\| e^{(t)} \|_2 
	\leq  \rho^t \| e^{(0)}\|_2 +  \frac{B_1}{\sqrt{N}}  + \alpha^2 B_2
	+ \frac{4}{\sqrt{N}}\frac{\beta\alpha \sqrt{t}\sqrt{K} [A(\delta/(3 Kt))]^{3}}{(1-\gamma)},
\end{eqnarray*}
where $e^{(t)}:= V^{(t)}-V$,  $\rho = (1-\beta)+\beta  [(1-\alpha)+\alpha\gamma]^K$, $B_1 ={\gamma} \frac{\Lambda \sqrt{| S|}   }{(1-\gamma)^2(1-[1-\alpha(1-\gamma)]^K)}$, $B_2 = \gamma^2 \frac{C\Delta \sqrt{|\mathcal{S}|}}{(1-\gamma)(1-[1-\alpha(1-\gamma)]^K)}$,    $A(\delta)=\sqrt{2(\log(1/\delta+1/4)}$,
$C = \exp(K(C_P+C_\mu))$
where $\| \hat P_i^l \|_2 \leq C_P$ and $\| P_\mu^l \|_2 \leq C_\mu$ for $l,C_P,C_\mu \geq 0$, and $\beta,\alpha\in(0,1)$.
\end{theorem}

\begin{remark}
\textsf{FedTD(0)} under i.i.d. sampling achieves high-probability convergence at the linear rate with the $\frac{B_1}{\sqrt{N}}$ error term due to the model mismatch $\Lambda$, the $\alpha^2 B_2$ error term due to the model mismatch $\Delta$, and the $\frac{4}{\sqrt{N}}\frac{\beta\alpha \sqrt{t}\sqrt{K} [A(\delta/(3Kt))]^{3}}{(1-\gamma)}$ error term due to i.i.d. sampling.
Unlike the single-agent case from Theorem~\ref{thm:single_iid}, \textsf{FedTD(0)}  achieves the $\sqrt{N}$-speedup, where $N$ is the number of agents.
\end{remark}

\begin{remark}
The step size $\alpha$, the federated parameter $\beta$, the number of local steps $K$, and the number of agents $N$ impact the convergence rate and residual error terms of \textsf{FedTD(0)} under i.i.d. sampling. 
We can reduce the error term due to i.i.d. sampling at the price of slow convergence, either by decreasing $\alpha$ and $\beta$. 
For instance, we can reduce the error term due to i.i.d. sampling by choosing $\beta = \frac{1}{T^\eta}$ and $\alpha = \frac{1}{K^\eta}$ with $\eta \in (1/2,1)$, which yields
\begin{eqnarray*}
 \frac{\beta\alpha \sqrt{t}\sqrt{K} [A(\delta/(3t))]^3}{(1-\gamma)}  = \mathcal{O}\left( \frac{\sqrt{t}(\log (Kt/\delta))^{3/2}}{T^{\eta}}  \frac{1}{K^{\eta-1/2}} \right).  
\end{eqnarray*}
Furthermore, the error term due to $\Delta$ and $\Lambda$ can be decreased only by reducing $\alpha$ and only by increasing the number of agents $N$, respectively. 
\end{remark}

\begin{remark}
Under i.i.d. sampling, \textsf{FedTD(0)} can be shown to achieve a significantly more accurate value function $V^{(t)}$ than single-agent \textsf{TD(0)}. We can show this by proving that \textsf{FedTD(0)} attains lower residual errors than single-agent \textsf{TD(0)}.     
First, two error terms due to $\Lambda$ and i.i.d. sampling of \textsf{FedTD(0)}, unlike single-agent \textsf{TD(0)},   vanish, as $N$ approaches $+\infty$. Second, the error term due to $\Delta$ of \textsf{FedTD(0)} can be proved to be lower than single-agent \textsf{TD(0)} by a factor of $1/(1-\gamma)$  by   setting $K\rightarrow+\infty$ and $\alpha = \sqrt{\frac{1}{\gamma C}}$ in Theorem~\ref{thm:multi_iid}. This yields
\begin{eqnarray*}
    \alpha^2\frac{\gamma^2  C\Delta \sqrt{|\mathcal{S}|}}{(1-\gamma)(1-[1-\alpha(1-\gamma)]^K)}
    \approx  \alpha^2\gamma^2 \frac{C\Delta \sqrt{|\mathcal{S}|}}{(1-\gamma)} =  \frac{\gamma\Delta \sqrt{|\mathcal{S}|}}{(1-\gamma)} \overset{\gamma \in (0,1)}{\leq} \frac{\gamma\Delta\sqrt{|\mathcal{S}|}}{(1-\gamma)^2}.
\end{eqnarray*}
In conclusion, as $N$ grows, \textsf{FedTD(0)} under i.i.d. sampling from Theorem~\ref{thm:multi_iid} achieves the lower residual error than  single-agent \textsf{TD(0)} from Theorem~\ref{thm:single_iid}.
\end{remark}

\subsection{Analysis for Markovian Setting} \label{sec:MA_Markov}
Finally, we establish the convergence of \textsf{FedTD(0)} under Markovian sampling. 
To show this, we define the mixing time in the federated setting as follows.

\begin{definition}\label{def:mixing_time_federated}
Let $\tau_{\epsilon}$ be the minimum time such that the following holds: For any agent, $k\geq0$ and $t\geq \tau_\epsilon$,
\(
    \| \xi_i^{(t,k)} \|_2 \leq \epsilon.
\)
Here, $\xi_i^{(t,k)}\in\R^{|\mathcal{S}|}$ is defined by
$\xi_i^{(t,k)}(s)=\delta_i^{(t,k)}-\mathbb{E}\bigl[\delta_i^{(t,k)} \mid \mathcal{F}^{(t,k)}\bigr]$ when $s=s_i^{(t,k)}$, and $\xi_i^{(t,k)}=0$ for all other $s$, where $\mathcal{F}^{(t,k)}$ is the filtration up to iterations $t,k$.
\end{definition}

We establish the convergence theorem of \textsf{FedTD(0)} under Markovian sampling, which achieves the same $\sqrt{N}$-speedup as \textsf{FedTD(0)} under i.i.d. sampling.

\begin{theorem}[Federated, Markovian sampling] \label{thm:multi_markov}
Consider \textsf{FedTD(0)} (Algorithm~\ref{alg:fed_TDlearning}) under i.i.d. sampling. Let each row-stochastic transition matrix $\hat P_i$ of the $i^{\rm th}$-agent  satisfy Assumption~\ref{assum:hat_P}. Then, 
	\begin{eqnarray*}
		\| e^{(t)} \|_2 
		& \leq & \rho^t \| e^{(0)}\|_2   +  \frac{B_1}{\sqrt{N}}   + \alpha^2 B_2 
 + \beta\frac{1}{1-\gamma}\left(\frac{2\tau_\beta}{1-\gamma} + t \beta \right),
	\end{eqnarray*}    
	where $e^{(t)}:= V^{(t)}-V$, 
 $\rho = (1-\beta)+\beta  [(1-\alpha)+\alpha\gamma]^K$, $B_1 ={\gamma} \frac{\Lambda \sqrt{| S|}   }{(1-\gamma)^2(1-[1-\alpha(1-\gamma)]^K)}$, $B_2 = \gamma^2 \frac{C\Delta \sqrt{|\mathcal{S}|}}{(1-\gamma)(1-[1-\alpha(1-\gamma)]^K)}$, $C = \exp(K(C_P+C_\mu))$ where $\| \hat P_i^l \|_2 \leq C_P$ and $\| P_\mu^l \|_2 \leq C_\mu$ for $l,C_P,C_\mu \geq 0$,   $\tau_\beta$ is defined by Definition~\ref{def:mixing_time_federated}, and $\beta,\alpha\in(0,1)$.
\end{theorem}
\begin{remark}
\textsf{FedTD(0)} under Markovian sampling, similar to i.i.d. sampling,  attains linear  convergence with the $\frac{B_1}{\sqrt{N}}$ error term due to the model mismatch $\Lambda$, the $\alpha^2 B_2$ error term due to the model mismatch $\Delta$, and the $\beta\frac{1}{1-\gamma}\left(\frac{2\tau_\beta}{1-\gamma} + t \beta \right)$ error term due to Markovian sampling $\tau_\beta$. 
Unlike \textsf{FedTD(0)} under i.i.d. sampling, which achieves the $\sqrt{N}$-speedup for two error terms due to $\Lambda$ and i.i.d. sampling, \textsf{FedTD(0)} under Markovian sampling attains the $\sqrt{N}$-speedup only for the  error term due to $\Lambda$.
\end{remark}

\begin{remark}
The step size $\alpha$, the federated parameter $\beta$, the number of local steps $K$, and the number of agents $N$ influence the convergence rate and residual error terms of \textsf{FedTD(0)} under Markovian sampling.
First, we can reduce the error term due to $\Delta$ at the price of slow convergence by decreasing $\alpha$.
Second, decreasing $\beta$ lessens the error term due to Markovian sampling $\tau_\beta$ at the cost of slow convergence. 
For instance, the error term due to Markovian sampling can be lessened by choosing $\beta = \frac{1}{T^\eta}$ with $\eta\in (1/2,1)$, as
\begin{eqnarray*}
    \beta\frac{1}{(1-\gamma)}\left(\frac{2\tau_\beta}{1-\gamma} + t \beta \right) = \mathcal{O}\left(\frac{1}{T^\eta} \right) + \mathcal{O}\left( \frac{1}{T^{2\eta-1}} \right). 
\end{eqnarray*}
Third, increasing the number of agents $N$ decreases the error term due to $\Lambda$. 
\end{remark}

\begin{remark}
Under Markovian sampling, like i.i.d. sampling, \textsf{FedTD(0)} can be shown to achieve higher accuracy of value functions $V^{(t)}$ than single-agent \textsf{TD(0)}.
We show this by proving that under certain conditions, two residual error terms of \textsf{FedTD(0)} due to $\Lambda$ and $\Delta$  from Theorem~\ref{thm:multi_markov} are lower than single-agent \textsf{TD(0)} from Theorem~\ref{thm:single_markov}. 
First, the error term due to $\Lambda$ for  \textsf{FedTD(0)} vanishes, as the number of agents $N$ goes to $+\infty$. 
Second, the error term due to $\Delta$ for \textsf{FedTD(0)} is lower than single-agent \textsf{TD(0)} by a factor of $1/(1-\gamma)$ for $\gamma \in (0,1)$ by choosing $K\rightarrow +\infty$ and $\alpha = \sqrt{\frac{1}{\gamma C}}$.
\end{remark}

\section{Experiments} \label{sec:Experiments}
\begin{figure*}[!htb] 
\centering
\includegraphics[width=0.49\textwidth]{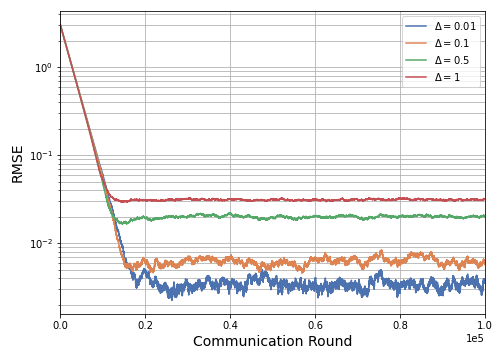}
\hfill
\includegraphics[width=0.49\textwidth]{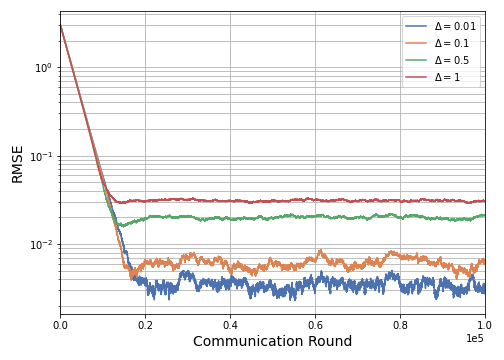}
\caption{Impact of the model mismatch $\Delta$ on the RMSE of the $V$-estimates for \textsf{FedTD(0)} with (\textbf{Left}) i.i.d sampling and (\textbf{Right}) Markovian sampling. Here, $N=10$, $K=5$, $\alpha=0.01$, and $\beta=0.4$.}
\label{fig:delta_iid&markov}
\end{figure*}
\subsection{Setup} \label{sec:Setup}
We evaluate the empirical performance of \textsf{FedTD(0)} under varying levels of model mismatch. 
We consider a randomly generated MDP with 10 states, where the transition matrix $P_\mu$ is row-stochastic and generated using uniform distributions. The reward function $r$ also has a uniform $[0,1]$ distribution. 
To simulate heterogeneity among agents, we introduce small perturbations to the transition matrix $P_\mu$, ensuring that each agent $i$ interacts with a slightly different transition model $\hat{P}_i$. The perturbation is controlled by a parameter $\Delta$, such that the Frobenius norm difference satisfies $\|\hat{P}_i - P_\mu\|_2 \leq \Delta$. This perturbation is followed by a projection step to ensure that each $\hat{P}_i$ remains row-stochastic. The reward function is kept identical across agents to isolate the effect of transition dynamics mismatch.
To quantify the error in value function estimation, we compute the root mean square error (RMSE) between the global value estimate $V^{(t)}$ and the true value function $V$, where $V$ is computed via the Bellman fixed-point equation $(I - \gamma P_\mu) V = r$. We track the evolution of this metric over communication rounds $t$. The value function is initialized with zero entries ($V^{(0)} = 0$ for all $s$), and we set $\gamma = 0.8$. All experiments are repeated with five different seed numbers, and results are averaged for robustness.

\subsection{Discussion} \label{sec:Discussion}
\begin{figure*}[!htb] 
\centering
\includegraphics[width=0.49\textwidth]{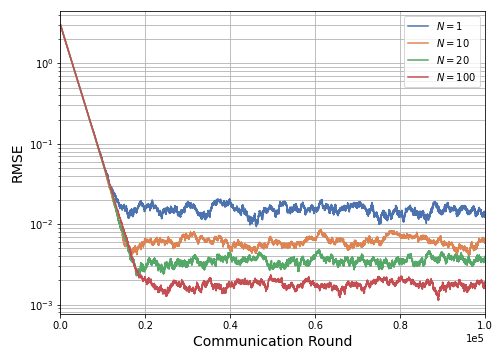}
\includegraphics[width=0.49\textwidth]{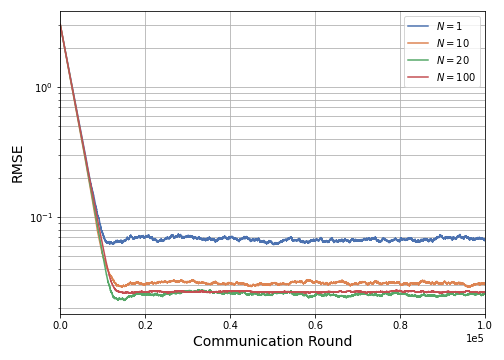}
\caption{Impact of the number of agents $N$ on the RMSE of $V$-estimates for \textsf{FedTD(0)} with Markovian sampling. Here, $K=5$, $\alpha=0.01$, and $\beta=0.4$. (\textbf{Left}) Corresponds to a model mismatch of $\Delta=0.1$, while (\textbf{Right}) considers a more severe mismatch of $\Delta=1$.}
\label{fig:N_markov}
\end{figure*}
\begin{figure*}[!htb]
\centering
    \includegraphics[width=0.49\textwidth]{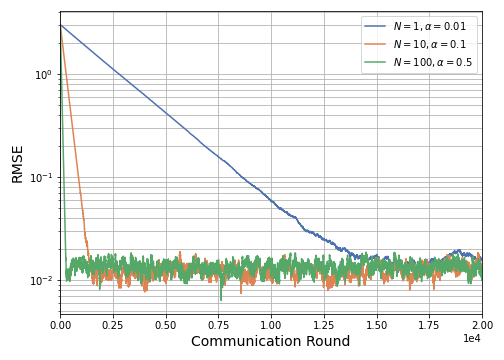}
    \label{N&alpha_markov}
    \hfill
    \includegraphics[width=0.49\textwidth]{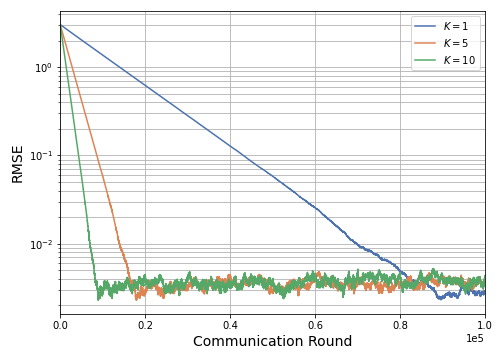}
    \label{K_markov}
\caption{RMSE of the $V$-estimates for \textsf{FedTD(0)} with Markovian sampling. (\textbf{Left}) Examines the effect of the number of agents $N$ and the learning rate $\alpha$ with $K=5$. (\textbf{Right}) Studies the effect of the number of local steps $K$ while keeping $N=20$ and $\alpha=0.01$. Both cases use $\Delta=0.1$ and $\beta=0.4$.}
    \label{fig:N&alpha_K_markov}
\end{figure*}
\begin{figure*}[!htb] 
\centering
\includegraphics[width=0.49\textwidth]{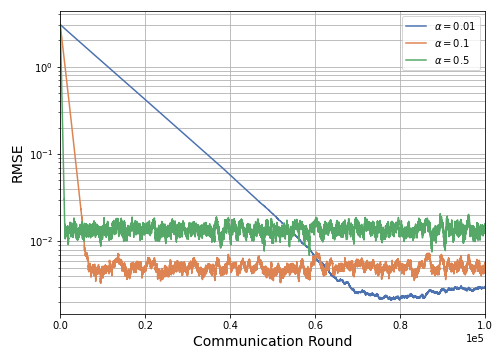}
\hfill
\includegraphics[width=0.49\textwidth]{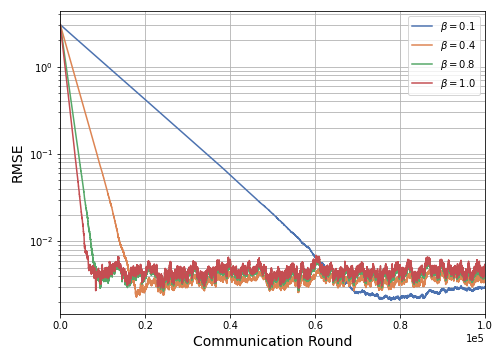}
\caption{RMSE of the $V$-estimates for \textsf{FedTD(0)} with Markovian sampling. (\textbf{Left}) Examines the effect of the learning rate $\alpha$ with $\beta=0.1$. (\textbf{Right}) Studies the effect of the federated parameter $\beta$ while $\alpha=0.01$. Both cases use $N=20$, $K=5$, and $\Delta=0.1$.}
\label{fig:alpha&beta_markov}
\end{figure*}

\paragraph{Effect of Model Mismatch under I.I.D. and Markovian Sampling.} 
The model mismatch $\Delta$ affects the accuracy of value function estimates under both i.i.d.\ and Markovian sampling. 
From Figure~\ref{fig:delta_iid&markov}, increasing $\Delta$ leads to a higher RMSE in the estimated value function. 
Interestingly, the residual error behaves almost identically in both sampling regimes, as further illustrated in Appendix~\ref{sec:AdditionalExperiments_IID}, Figure~\ref{figAppendix:delta_iid} and~Appendix~\ref{sec:AdditionalExperiments_Markov}, Figure~\ref{figAppendix:delta_markov}.
This confirms our theoretical analysis in Theorems \ref{thm:multi_iid} and \ref{thm:multi_markov}, which state that while the convergence dynamics differ under i.i.d.\ and Markovian sampling, the final bias due to model mismatch is primarily governed by $\Delta$. 
In particular, a large value of $\Delta$ results in a large residual error around the unique fixed value function $V$. 
These results highlight that FedRL methods must account for model mismatch effects rather than focusing solely on the sampling strategy of individual agents.

\paragraph{Effect of the Number of Agents on Reducing Model Mismatch Bias.}
From Figure~\ref{fig:N_markov}, \textsf{FedTD(0)} with the larger number of agents $N$ ensures that its estimated value function is closer to the true value function. This corroborates the $\sqrt{N}$-speedup in the convergence in Theorems \ref{thm:multi_iid} and \ref{thm:multi_markov}.
Further validation of this trend is provided in Appendix~\ref{sec:AdditionalExperiments_Delta}, Figure~\ref{figAppendix:N_markov}.

\paragraph{Convergence Speedup with More Agents.} 
As shown in Figure~\ref{fig:N&alpha_K_markov}, more agents allow for a larger step size $\alpha$ without destabilizing the learning process. This is because federated averaging reduces variance in updates, enabling more aggressive updates without divergence. Consequently, \textsf{FedTD(0)} with more agents achieves similar accuracy in fewer iterations compared to a single-agent setting, making it a scalable solution for distributed RL applications. This aligns with the $\sqrt{N}$-speedup and the lower residual error by multiple agents from Theorems \ref{thm:multi_iid} and \ref{thm:multi_markov}.

\paragraph{Robustness to the Choice of Local Steps $K$: Communication Efficiency.} 
As shown in Figure~\ref{fig:N&alpha_K_markov}, increasing $K$ reduces communication overhead without significantly compromising final performance. Specifically, $K=10$ achieves the same RMSE as $K=1$, but with a \emph{10-fold reduction in communication rounds}. 
This suggests that increasing local updates can substantially improve efficiency in FedRL, making it particularly useful in scenarios where the communication cost is expensive.

\paragraph{The Effect of Learning Rate $\alpha$ and Federated Parameter $\beta$.} 
Figure~\ref{fig:alpha&beta_markov} illustrates how $\alpha$ and $\beta$ impact the performance of \textsf{FedTD(0)}. A smaller $\alpha$ slows convergence but stabilizes learning, whereas a larger $\alpha$ leads to faster convergence. Similarly, a large $\beta$ allows faster adaptation of the global value estimate. However, changes in $\alpha$ have a greater effect on residual error compared to adjustments in $\beta$, as supported by our non-asymptotic results (see Theorem \ref{thm:multi_markov}), where the convergence of \textsf{FedTD(0)} consists of the two residual error terms, $\mathcal{O}(\alpha^2)+\mathcal{O}(\beta)$.

\section{Conclusion} \label{sec:Conclusion}
In this paper, we investigated \textsf{FedTD(0)} for \emph{policy evaluation} under \emph{model mismatch}, where multiple agents interact with perturbed environments and \emph{periodically} exchange value estimates. We established \emph{linear convergence} guarantees for single-agent \textsf{TD(0)} under both \emph{i.i.d.} and \emph{Markovian} sampling regimes and demonstrated how environmental perturbations introduce systematic bias in individual learning. Extending these results to the federated setting, we quantified the role of model mismatch, network connectivity, and mixing behavior in the convergence of \textsf{FedTD(0)}.
Our theoretical results indicate that even under heterogeneous transition dynamics, moderate levels of information sharing among agents can effectively mitigate environment-specific errors and improve convergence rates.
Empirical results further support these findings, demonstrating that federated collaboration reduces individual bias and accelerates convergence to the true value function.
These findings highlight the potential of FedRL for real-world applications where identical environment assumptions are impractical, such as multi-robot coordination and decentralized control in sensor networks. 
For future research, we aim to extend our framework beyond policy evaluation to control settings, such as federated $Q$-learning, and explore the effectiveness of FedRL in real-world tasks.

\section*{Acknowledgments} \label{sec:Ack}
This work was partially supported by the Swedish Research Council through grant agreement no. 2024-04058 and in part by Sweden's Innovation Agency (Vinnova). The computations were enabled by resources provided by the National Academic Infrastructure for Supercomputing in Sweden (NAISS) at Chalmers Centre for Computational Science and Engineering (C3SE) partially 
funded by the Swedish Research Council through grant agreement no. 2022-06725. 
The research reported in this publication was supported by funding from King Abdullah University of Science and Technology (KAUST): i) KAUST Baseline Research Scheme, ii) Center of Excellence for Generative AI, under award number 5940, iii) SDAIA-KAUST Center of Excellence in Artificial Intelligence and Data Science.

\section*{Supplementary Material} \label{sec:SI}
For comprehensive supplementary appendices and accompanying code, readers are directed to the full version of this paper, accessible via Springer Link and arXiv \cite{beikmohammadi2025collaborative}, along with the corresponding code repository hosted on GitHub: \url{https://github.com/AliBeikmohammadi/FedRL}.

\bibliographystyle{splncs04}
\bibliography{ref}

\clearpage
\appendix 

\newpage
\section{Notation} \label{sec:Notation}
Let $\mathbb{N}$, $\mathbb{N}_0$, and $\mathbb{R}$ denote the positive integers, non-negative integers, and real numbers, respectively.  
For \(a,b \in \mathbb{N}_0\) with \(a \le b\), write 
\(\,[a,b] = \{\,a,\,a+1,\,\ldots,\,b\}\).
For any vector \(v = (v(1),v(2),\ldots, v(d)) \in \mathbb{R}^d\), the $\ell_1$-, $\ell_2$-, and $\ell_\infty$-norms are:
\[
\|v\|_1 \;\coloneqq\; \sum_{i=1}^d |v(i)|,\quad
\|v\|_2 \;\coloneqq\; \sqrt{\sum_{i=1}^d \big(v(i)\big)^2},\quad
\|v\|_\infty \;\coloneqq\; \max_{1 \le i \le d} \bigl|v(i)\bigr|.
\]

\section{Basic Inequalities}
We present basic inequalities that are useful for our convergence analysis. First, Lemma~\ref{lemma:recursion_trick} presents the explicit expression of $e^{(t)}$ governed by the recursion $e^{(t+1)} = A e^{(t+1)} + B^{(t)}$, while Lemma~\ref{lemma:vector_bernstein} introduces the vector Bernstein inequality derived by \cite{kohler2017sub}.

\begin{lemma}\label{lemma:recursion_trick}
Let $e^{(t+1)} = A e^{(t+1)} + B^{(t)}$ for $t \geq 0$. Then, $e^{(t)}=A^t e^{(0)}+\sum_{l=0}^{t-1} A^{t-1-l} B^{(l)}$.
\end{lemma}

\begin{lemma}(Lemma 18 of \cite{kohler2017sub})\label{lemma:vector_bernstein}
Let $x_1,\ldots,x_n$ be independent random vectors in $\mathbb{R}^d$ that satisfy
\begin{eqnarray*}
    \mathbb{E}[x_i]=0,\quad \| x_i \|_2 \leq \mu, \quad \text{and} \quad \mathbb{E}\| x_i \|_2^2 \leq \sigma^2.
\end{eqnarray*}
Then, for $0< \epsilon \leq \sigma^2/\mu$, 
\begin{eqnarray*}
    \text{\rm Pr}\left( \left\| \frac{1}{n}\sum_{i=1}^n x_i \right\| \geq \epsilon \right) \leq \exp\left( - n \cdot \frac{\epsilon^2}{8\sigma^2}+\frac{1}{4} \right),
\end{eqnarray*}
or equivalently with probability at least $1-\delta$ 
\begin{eqnarray*}
    \left\| \frac{1}{n}\sum_{i=1}^n x_i \right\| \leq \sqrt{\frac{8\sigma^2}{n}\left( \log(1/\delta) + 1/4\right)}.
\end{eqnarray*}
\end{lemma}

Next, the following lemmas are useful for analyzing single-agent \textsf{TD(0)} algorithms. In particular, 
Lemma~\ref{lemma:e_norm_D} presents the recursion of $ V^{(t+1)} - V$, while Lemma~\ref{lemma:boundedness_of_xi} bounds $| \delta^{(t)} - \mathbb{E}[\delta^{(t)}| \mathcal{F}^{(t)}] |$, where  $\delta^{(t)}=r^{(t)} + \gamma V^{(t)}(s^{(t+1)})-V^{(t)}(s^{(t)})$.

\begin{lemma}\label{lemma:e_norm_D}
Consider the single-agent \textsf{TD(0)} algorithm (Algorithm~\ref{alg:single_agent_TDlearning}). 
Let a row-stochastic transition matrix $\hat P$ satisfy Assumption~\ref{assum:hat_P_single}. Then, 
\begin{eqnarray*}
 e^{(t+1)} 
= [(1-\alpha)I  + \alpha \gamma \hat P]e^{(t)} + \alpha\gamma (\hat P - P_\mu)V + \alpha\xi^{(t)},
\end{eqnarray*}
where  $e^{(t)} := V^{(t)}-V$, and $\xi^{(t)} \in \R^{|\mathcal{S}|}$ is defined by
\[
   \xi^{(t)}(s) \;=\; 
   \begin{cases}
   \delta^{(t)} \;-\;\mathbb{E}\bigl[\delta^{(t)} \mid \mathcal{F}^{(t)}\bigr], & s = s^{(t)}, \\[4pt]
   0, & \text{otherwise}.
   \end{cases}
\]
\end{lemma}
Here, $\delta^{(t)} = r^{(t)} + \gamma V^{(t)}(s^{(t+1)})-V^{(t)}(s^{(t)})$ and $\mathcal{F}^{(t)}$ is the filtration up to step $t$. 

\begin{lemma}\label{lemma:boundedness_of_xi}
Let $V^{(t+1)} = V^{(t)} + \alpha \delta^{(t)}$ where $\delta^{(t)} = r^{(t)} + \gamma V^{(t)}(s^{(t+1)})-V^{(t)}(s^{(t)})$, and $V = R + \gamma P_\mu V$. If  $\gamma\in (0,1)$,  $r^{(t)}\in [0,1]$, and $V^{(0)}=0$, then 
\begin{eqnarray*}
    V(s) \in [0,M], \quad \text{and} \quad V^{(t)}(s) \in [0,M],
\end{eqnarray*}    
where $M:= 1/(1-\gamma)$. Furthermore, 
\[
\bigl|\delta^{(t)}\bigr|  \leq M, \quad \text{and}    \quad   \Bigl|\delta^{(t)} - \mathbb{E}[\delta^{(t)} \mid \mathcal{F}^{(t)}]\Bigr| \leq 2M.
\]
\end{lemma}

Finally, we present the following lemma for establishing the linear convergence of \textsf{FedTD(0)} under i.i.d. sampling (in high probability) and Markovian sampling in Sections~\ref{sec:MA_IID} and~\ref{sec:MA_Markov}, respectively.

\begin{lemma}\label{lemma:multi_aget_TD}
Consider \textsf{FedTD(0)} (Algorithm~\ref{alg:fed_TDlearning}). Let each row-stochastic transition matrix $\hat P_i$ of the $i^{\rm th}$-agent  satisfy Assumption~\ref{assum:hat_P}. Then, 
\begin{eqnarray*}
e^{(t+1)} 
& = &  \hat A e^{(t)} +  \beta\alpha\gamma \cdot Y + \beta\alpha  \cdot Z^{(t)}, 
\end{eqnarray*}
where $e^{(t)}:=V^{(t)}-V$,  $\hat A = (1-\beta)I + \frac{\beta}{N}\sum_{i=1}^N A_i^K$, $Y=\frac{1}{N}\sum_{i=1}^N \sum_{l=0}^{K-1}A_i^{K-1-l} (\hat P_i-P_{\mu})V$,  $Z^{(t)} = \frac{1}{N}\sum_{i=1}^N\sum_{l=0}^{K-1}A_i^{K-1-l} \xi_i^{(t,l)}$, $A_i = (1-\alpha)I + \alpha\gamma \hat P_i$, $\xi_i^{(t,k)} \in \mathbb{R}^{|\mathcal{S}|}$ is defined by 
\[
   \xi_i^{(t,k)}(s) \;=\; 
   \begin{cases}
   \delta_i^{(t,k)} \;-\;\mathbb{E}\bigl[\delta_i^{(t,k)} \mid \mathcal{F}^{(t,k)}\bigr], & s = s_i^{(t,k)}, \\[4pt]
   0, & \text{otherwise}
   \end{cases}
\]
and $\mathcal{F}^{(t,k)}$ is the filtration up to iterations $t,k$.
\end{lemma}

\section{Proof of Main Results}
In this section, we provide the complete proofs for all primary lemmas and theorems. 
\subsection{Proof of Lemma~\ref{lemma:e_norm_D}}
\begin{proof}
We prove the results in the following steps. 
\paragraph{Step 1) Derive $\mathbb{E}[\delta^{(t)} \mid \mathcal{F}^{(t)}]$.}  Let $\delta^{(t)}=r^{(t)} + \gamma V^{(t)}(s^{(t+1)})-V^{(t)}(s^{(t)})$, and $\mathcal{F}^{(t)}$ be the filtration up to step $t$.
Furthermore, denote $(\hat T V)(s) = R(s) + \gamma \sum_{s'} \hat P(s'|s) V(s')$ and $(T V)(s) = R(s) + \gamma \sum_{s'}  P_\mu(s'|s) V(s')$.
Then, 
\begin{eqnarray*}
\mathbb{E}[\delta^{(t)} \mid \mathcal{F}^{(t)}] 
&=& \big(\hat T V^{(t)}\big)(s^{(t)}) - V^{(t)}(s^{(t)}) \\
&=& \big(\hat T V^{(t)}\big)(s^{(t)}) - \big(T V^{(t)}\big)(s^{(t)}) + \big(T V^{(t)}\big)(s^{(t)})  - V^{(t)}(s^{(t)}) \\
& = & \gamma \sum_{s'} [\hat P(s'|s^{(t)}) - P_\mu(s'|s^{(t)})] V^{(t)}(s')+ \big(T V^{(t)}\big)(s^{(t)})  - V^{(t)}(s^{(t)}). 
\end{eqnarray*}

\paragraph{Step 2) Derive the recursion of  $e^{(t)}:= V^{(t)} - V$.}
Let $e^{(t)}(s^{(t)}):= V^{(t)}(s^{(t)}) - V(s^{(t)})$. Then, 
\begin{eqnarray*}
    e^{(t+1)}(s^{(t)})
    & = & V^{(t+1)}(s^{(t)}) - V(s^{(t)}) \\ 
    & \overset{V^{(t+1)}(s^{(t)})}{=} & V^{(t)}(s^{(t)}) -   V(s^{(t)}) + \alpha \delta^{(t)} \\
    & \overset{e^{(t)}(s^{(t)})}{=} & e^{(t)}(s^{(t)}) + \alpha \delta^{(t)} \\
    & = & e^{(t)}(s^{(t)}) + \alpha \mathbb{E}[\delta^{(t)}| \mathcal{F}^{(t)}] + \alpha[\delta^{(t)} - \mathbb{E}[\delta^{(t)}| \mathcal{F}^{(t)}]]   \\
    & \overset{\mathbb{E}[\delta^{(t)}| \mathcal{F}^{(t)}]}{=} & e^{(t)}(s^{(t)}) + \alpha[\big(T V^{(t)}\big)(s^{(t)}) - V^{(t)}(s^{(t)})] \\
    && + \alpha \gamma \sum_{s'} [\hat P(s'|s^{(t)}) - P_\mu(s'|s^{(t)})] V^{(t)}(s') + \alpha[\delta^{(t)} - \mathbb{E}[\delta^{(t)}| \mathcal{F}^{(t)}]].
\end{eqnarray*}

Next, since $V^{(t)}(s')=V(s')+e^{(t)}(s')$,
\begin{eqnarray*}
    e^{(t+1)}(s^{(t)})
     & = & e^{(t)}(s^{(t)}) + \alpha \gamma \sum_{s'} [\hat P(s'|s^{(t)}) - P_\mu(s'|s^{(t)})] e^{(t)}(s') \\ 
     && + \alpha[\big(T V^{(t)}\big)(s^{(t)}) - V^{(t)}(s^{(t)})] \\
    &&  + \alpha \gamma \sum_{s'} [\hat P(s'|s^{(t)}) - P_\mu(s'|s^{(t)})] V(s') + \alpha[\delta^{(t)} - \mathbb{E}[\delta^{(t)}| \mathcal{F}^{(t)}]].
\end{eqnarray*}

Next, let $e^{(t)}$ and $V^{(t)}$ be the vector in $\R^{|\mathcal{S}|}$ with its element $e^{(t)}(s)$ and $V^{(t)}(s)$, respectively, for $s\in S$. Then, 
\[
  e^{(t+1)} = e^{(t)} + \alpha \gamma [\hat P - P_\mu] e^{(t)}+ \alpha[T V^{(t)} - V^{(t)}] + \alpha\gamma (\hat P - P_\mu)V + \alpha\xi^{(t)},
\]
where $\xi^{(t)} \in \R^{|\mathcal{S}|}$ is defined by
\[
   \xi^{(t)}(s) \;=\; 
   \begin{cases}
   \delta^{(t)} \;-\;\mathbb{E}\bigl[\delta^{(t)} \mid \mathcal{F}^{(t)}\bigr], & s = s^{(t)}, \\[4pt]
   0, & \text{otherwise}.
   \end{cases}
\]

\paragraph{Step 3) Derive $T V^{(t)} - V^{(t)}$.} 
Recall that $T$ is defined by 
\[
TV^{(t)} \;=\; R \;+\;\gamma\,P_\mu V^{(t)},
\]
where $R$ is the vector in $\R^{|\mathcal{S}|}$ with its element $r(s)$ and $V$ satisfies 
\[
V \;=\; TV \;=\; R + \gamma\,P_\mu V.
\]
Then,  
\[
\begin{aligned}
T\,V^{(t)} - V^{(t)}
&=\; \bigl(R + \gamma\,P_\mu\,V^{(t)}\bigr) - \bigl(V + e^{(t)}\bigr)
\\[6pt]
&\overset{(\star)}{=}\; R + \gamma\,P_\mu\bigl(V + e^{(t)}\bigr) \;-\; V \;-\; e^{(t)}
\\[6pt]
& \overset{(\star\star)}{=}\; \gamma\,P_\mu\,e^{(t)} \;-\; e^{(t)},
\end{aligned}
\]
where we reach $(\star)$ by the fact that $V^{(t)} = V+e^{(t)}$, and $(\star\star)$ by the fact that $V=R+\gamma P_\mu V$. Therefore, plugging the expression of $T\,V^{(t)} - V^{(t)}$ into the recursion of $e^{(t+1)}$, we obtain
\begin{eqnarray*}
e^{(t+1)} 
&=&  [(1-\alpha)I -\alpha\gamma (P_\mu-\hat P) + \alpha \gamma P_\mu]e^{(t)} + \alpha\gamma (\hat P - P_\mu)V + \alpha\xi^{(t)} \\
&=&  [(1-\alpha)I  + \alpha \gamma \hat P]e^{(t)} + \alpha\gamma (\hat P - P_\mu)V + \alpha\xi^{(t)}.
\end{eqnarray*}
We complete the proof. 
\end{proof}

\subsection{Proof of Lemma~\ref{lemma:boundedness_of_xi}}
\begin{proof}
We prove two statements as follows. 

\paragraph{The first statement.}	
We prove the first statement by induction. 
Let $\gamma\in (0,1)$, and the reward function $r^{(t)}=r(s^{(t)}) \in [0,1]$. 
If $V^{(0)}(s) \in \left[0, 1/(1-\gamma) \right]$, and $V^{(t)}(s) \in [0,1/(1-\gamma)]$,  then
\begin{eqnarray*}
	V^{(t+1)}(s^{(t)}) 
	& = & (1-\alpha) V^{(t)}(s^{(t)}) + \alpha[ r^{(t)} + \gamma V^{(t)}(s^{(t+1)})]  \\
	& \leq & (1-\alpha) \| V^{(t)} \|_\infty + \alpha[1 + \gamma \| V^{(t)} \|_\infty ] \\
	& \leq & \frac{1}{1-\gamma}.
\end{eqnarray*}	
Furthermore, since $V^{(0)}=0$ and $r^{(t)} \geq 0$, we can show that $V^{(1)}(s^{(t)}) \geq 0$ and therefore,  
	\begin{eqnarray*}
	V^{(t+1)}(s^{(t)}) 
	& = & (1-\alpha) V^{(t)}(s^{(t)}) + \alpha[ r^{(t)} + \gamma V^{(t)}(s^{(t+1)})]  \\
	& \geq & 0.
\end{eqnarray*}	
Hence, $V^{(t+1)}(s^{(t)}) \in [0,1/(1-\gamma)]$. Next, by the fact that $R \in [0,1]$ and $P_\mu$ is the row-stochastic matrix, i.e. $\| P_\mu \|_1 \leq 1$, we can show that 
\begin{eqnarray*}
	\| V \|_\infty 
	& \leq & \| R \|_\infty + \gamma \| P_\mu\|_1 \| V \|_\infty \\
	& \leq & 1 + \gamma \frac{1}{1-\gamma} = \frac{1}{1-\gamma},
\end{eqnarray*}	
and 
\begin{eqnarray*}
	V 
	& \geq  & \gamma P_\mu V \geq 0.
\end{eqnarray*}	
In conclusion, we obtain $V^{(t)}(s) \in [0,1/(1-\gamma)]$ and $V(s) \in [0,1/(1-\gamma)]$.

\paragraph{The second statement.}	
We prove the second statement from the first statement. 
By the definition of $\delta^{(t)}$, and  the fact that  $V^{(t)}(s)\in [0,1/(1-\gamma)]$ and $r^{(t)} \in [0,1]$, we can show that $\delta^{(t)} \in \left[ - \frac{1}{1-\gamma} , \frac{1}{1-\gamma} \right]$, and $\Bigl|\delta^{(t)} - \mathbb{E}[\delta^{(t)} \mid \mathcal{F}^{(t)}]\Bigr| \leq \frac{2}{1-\gamma}]$.
We complete the proof. 
\end{proof}

\subsection{Proof of Lemma~\ref{lemma:multi_aget_TD}}
\begin{proof}
Define $e_i^{(t,K)}:=V_i^{(t,K)}-V$, and $e^{(t)} = V^{(t)}-V$. Then, by following the proof arguments in Lemma~\ref{lemma:e_norm_D}, 
\begin{eqnarray*}
    e_i^{(t,k+1)} = A_i e_i^{(t,k)} + \alpha\gamma (\hat P_i-P_{\mu})V +\alpha \xi_i^{(t,k)},
\end{eqnarray*}
where $A_i = (1-\alpha)I + \alpha\gamma \hat P_i$,  $\xi_i^{(t,k)} \in \mathbb{R}^{|\mathcal{S}|}$ is defined by 
\[
   \xi_i^{(t,k)}(s) \;=\; 
   \begin{cases}
   \delta_i^{(t,k)} \;-\;\mathbb{E}\bigl[\delta_i^{(t,k)} \mid \mathcal{F}^{(t,k)}\bigr], & s = s_i^{(t,k)}, \\[4pt]
   0, & \text{otherwise}
   \end{cases}
\]
and $\mathcal{F}^{(t,k)}$ is the filtration up to iterations $t,k$.

Next, by applying the equation recursively over $k=0,1,\ldots,K-1$,
\begin{eqnarray*}
    e_i^{(t,K)} 
    & =&  A_i^K e_i^{(t,0)} + \alpha\gamma \sum_{l=0}^{K-1}A_i^{K-1-l} (\hat P_i-P_{\mu})V +\alpha \sum_{l=0}^{K-1}A_i^{K-1-l} \xi_i^{(t,l)} \\
    & \overset{V_i^{(t,0)}=V^{(t)}}{=} & A_i^K e^{(t)} + \alpha\gamma \sum_{l=0}^{K-1}A_i^{K-1-l} (\hat P_i-P_{\mu})V +\alpha \sum_{l=0}^{K-1}A_i^{K-1-l} \xi_i^{(t,l)}.
\end{eqnarray*}

Finally, from the definition of $e^{(t)}$, 
\begin{eqnarray*}
e^{(t+1)} 
& = & (1-\beta) e^{(t)} + \frac{\beta}{N}\sum_{i=1}^N e_i^{(t,K)} \\
& = &  \hat A e^{(t)} +  \beta\alpha\gamma \cdot Y + \beta\alpha  \cdot Z^{(t)}, 
\end{eqnarray*}
where $\hat A = (1-\beta)I + \frac{\beta}{N}\sum_{i=1}^N A_i^K$, $Y=\frac{1}{N}\sum_{i=1}^N \sum_{l=0}^{K-1}A_i^{K-1-l} (\hat P_i-P_{\mu})V$, and $Z^{(t)} = \frac{1}{N}\sum_{i=1}^N\sum_{l=0}^{K-1}A_i^{K-1-l} \xi_i^{(t,l)}$. We complete the proof.
\end{proof}

\subsection{Proof of Theorem~\ref{thm:single_iid}}
\begin{proof}
We prove the results in the following steps. 

\paragraph{Step 1) Bound $\| e^{(t)} \|_2$.}
From Lemma~\ref{lemma:e_norm_D}, 
\begin{eqnarray*}
e^{(t+1)} 
&=&  Ae^{(t)} + \alpha\gamma B + \alpha\xi^{(t)},
\end{eqnarray*}
where $A= (1-\alpha)I + \alpha \gamma \hat P$ and $B=(\hat P - P_\mu)V$. 
By using Lemma~\ref{lemma:recursion_trick},
\begin{eqnarray*}
    e^{(t)} & = & A^t e^{(0)} + \alpha\gamma \sum_{l=0}^{t-1} A^{t-1-l} B + \alpha \sum_{l=0}^{t-1} A^{t-1-l} \xi^{(l)}.
\end{eqnarray*}
From the definition of the $\ell_2$-norm, and by the triangle inequality, 
\begin{eqnarray*}
    \| e^{(t)} \|_2 
    & \leq & \| A^t\|_2 \|  e^{(0)} \|_2 + \alpha\gamma  \left\|  \sum_{l=0}^{t-1} A^{t-1-l} B \right\|_2  + \alpha \left\| \sum_{l=0}^{t-1}  Z^l\right\|_2 \\
    & \leq &  \| A\|_2^t \|  e^{(0)} \|_2 + \alpha\gamma  \left\|  \sum_{l=0}^{t-1} A^{t-1-l} B \right\|_2  + \alpha \left\| \sum_{l=0}^{t-1}  Z^l\right\|_2,
\end{eqnarray*}
where $Z^l =A^{t-1-l} \xi^{(l)}$. Next, since $\hat P$ is a row-stochastic matrix, we can prove that its maximum eigenvalue is upper-bounded by $1$, and that $\| A \|^t_2 \leq [(1-\alpha)+\alpha\gamma]^t$ for $t \geq 0$. Hence, 
\begin{eqnarray*}
    \| e^{(t)} \|_2 
    & \leq &  \rho^t \|  e^{(0)} \|_2 + \alpha\gamma  \left\|  \sum_{l=0}^{t-1} A^{t-1-l} B \right\|_2  + \alpha \left\| \sum_{l=0}^{t-1}  Z^l\right\|_2,
\end{eqnarray*}
where $\rho = (1-\alpha)+\alpha\gamma \in (0,1)$ for any $\alpha\in(0,1)$.

\paragraph{Step 2) Bound $\left\|  \sum_{l=0}^{t-1} A^{t-1-l} B \right\|_2 $.}
From Assumption~\ref{assum:hat_P_single}, 
\begin{eqnarray*}
	\left\|  \sum_{l=0}^{t-1-l} A^{t-1-l}  B \right\|_2
	& \leq & \sum_{l=0}^{t-1-l} \| A \|^{t-1-l}_2 \| B \|_2 \\
	& \leq & \sum_{l=0}^{t-1-l} \rho^{t-1-l} \| \hat P - P_\mu \|_2 \| V \|_2 \\
	& \leq & \Delta \sum_{l=0}^{t-1-l} \rho^{t-1-l}  \| V \|_2.
\end{eqnarray*}	
Since 
\begin{eqnarray*}
	\| V \|_2 & \overset{\text{Lemma~\ref{lemma:boundedness_of_xi}}}{\leq} & \sqrt{ |\mathcal{S}|  } M,  
\end{eqnarray*}	
where $|\mathcal{S}|$ is the number of states, we have 
\begin{eqnarray*}
		\left\|  \sum_{l=0}^{t-1-l} A^{t-1-l}  B \right\|_2
		& \leq &  \Delta \sqrt{ |\mathcal{S}|  } M \sum_{l=0}^{t-1-l} \rho^{t-1-l} \\
		& \leq & \Delta \sqrt{ |\mathcal{S}|  } M \sum_{l=0}^{\infty} \rho^{l} \\
		& = & \frac{\Delta \sqrt{|\mathcal{S}|} M }{\alpha(1-\gamma)}
\end{eqnarray*}	

\paragraph{Step 3) Bound $\left\| \sum_{l=0}^{t-1}  Z^l\right\|_2$ by the vector Bernstein inequality.} 
Under the i.i.d. sampling regime, and from Lemma~\ref{lemma:boundedness_of_xi}, we have $\mathbb{E}[\xi^{(l)}] =0$ and $\| \xi^{(l)} \|_2 \leq 2M$ with $M=1/(1-\gamma)$. 
Therefore, $Z^l =A^{t-1-l} \xi^{(l)}$  satisfies $\mathbb{E}[Z^l] = A^{t-1-l} \mathbb{E}[\xi^{(l)}] =  0$, $\| Z^l \|_2 \leq 2M$, and $\mathbb{E}\| Z^l \|_2^2 \leq 4 M^2$. Therefore, from Lemma~\ref{lemma:vector_bernstein}, 
\begin{eqnarray*}
    \left\|  \sum_{l=0}^{t-1} Z^l \right\|_2 \leq \sqrt{t}\sqrt{32 M^2(\log(1/\delta_1)+1/4)}
\end{eqnarray*}
with probability at least $1-\delta_1$. 

\paragraph{Step 4) Derive the high-probability convergence in $\| e^{(t)}\|_2$.}
By the concentration bounds for $\left\|  \sum_{l=0}^{t-1} A^{t-1-l} B \right\|_2 $ and $\left\| \sum_{l=0}^{t-1}  Z^l\right\|_2$, we have 
\begin{eqnarray*}
    \| e^{(t)} \|_2 
    & \leq &  \rho^t \|  e^{(0)} \|_2 + \gamma  \frac{\Delta \sqrt{|\mathcal{S}|} M }{(1-\gamma)}  + \alpha \sqrt{t}\sqrt{32 M^2(\log(t/\delta)+1/4)}, 
\end{eqnarray*}
with probability at least $1-\delta$ (as $\delta_1=\delta/t$).
We complete the proof. 
\end{proof}

\subsection{Proof of Theorem~\ref{thm:single_markov}}
\begin{proof}
By following Steps 1) and 2) in Theorem~\ref{thm:single_iid}, we obtain
\begin{itemize}
\item \( 
    \| e^{(t)} \|_2 
     \leq   \rho^t \|  e^{(0)} \|_2 + \alpha\gamma  \left\|  \sum_{l=0}^{t-1} A^{t-1-l} B \right\|_2  + \alpha \left\| \sum_{l=0}^{t-1}  Z^l\right\|_2,
\)
where $\rho = (1-\alpha)+\alpha\gamma \in (0,1)$ for any
$\alpha\in(0,1)$, $A=(1-\alpha)I+\alpha\gamma\hat P$, $B=(\hat P-P_\mu)V$, and $Z^l=A^{t-1-l}\xi^{(l)}$.
\item
\(
	\left\|  \sum_{l=0}^{t-1-l} A^{t-1-l}  B \right\|_2
	\leq \frac{\Delta \sqrt{|\mathcal{S}|} M }{\alpha(1-\gamma)},
\) where $|\mathcal{S}|$ is the number of states.
\end{itemize}

To derive the convergence bound under Markovian sampling, we must bound $\left\| \sum_{l=0}^{t-1}  Z^l\right\|_2$.
From the definition of the mixing time $\tau_{\epsilon}$,
\begin{eqnarray*}
    \left\| \sum_{l=0}^{t-1}  Z^l\right\|_2
    & \leq & \sum_{l=0}^{\tau_{\epsilon}-1} \| Z^l\|_2 + \sum_{l=\tau_\epsilon}^{t-1} \| Z^l \|_2.
\end{eqnarray*}
Since 
\begin{eqnarray*}
    \| Z^l\|_2 \leq \| A\|_2^{t-1-l} \| \xi^{(l)} \|_2 \leq \rho^{t-1-l} \| \xi^{(l)} \|_2, 
\end{eqnarray*}
we have 
\begin{eqnarray*}
    \left\| \sum_{l=0}^{t-1}  Z^l\right\|_2
    & \leq & \sum_{l=0}^{\tau_{\epsilon}-1} \rho^{t-1-l} \| \xi^{(l)} \|_2  + \sum_{l=\tau_\epsilon}^{t-1} \rho^{t-1-l} \| \xi^{(l)} \|_2 \\
    & \leq & \sum_{l=0}^{\tau_{\epsilon}-1} \rho^{t-1-l} \cdot 2M + \sum_{l=\tau_\epsilon}^{t-1} \rho^{t-1-l}\epsilon \\
    & \leq & \sum_{l=0}^{\tau_{\epsilon}-1} 2M + \sum_{l=0}^{\infty} \rho^{l}\epsilon \\
    & = & 2M \tau_{\epsilon} + \frac{\epsilon}{1-\rho}.
\end{eqnarray*}

Next, by plugging the worst-case bounds for $\left\|  \sum_{l=0}^{t-1} A^{t-1-l} B \right\|_2 $ and $\left\| \sum_{l=0}^{t-1}  Z^l\right\|_2$, 
\begin{eqnarray*}
\| e^{(t)} \|_2 
    & \leq &  \rho^t \|  e^{(0)} \|_2 + \gamma \frac{\Delta \sqrt{|\mathcal{S}|} M }{(1-\gamma)} + \alpha \left(2M \tau_{\epsilon} + \frac{\epsilon}{1-\rho} \right).
\end{eqnarray*}
By the definition of $\rho$ and by choosing $\epsilon=\alpha$, we obtain the final result. 
\end{proof}

\subsection{Proof of Theorem~\ref{thm:multi_iid}}
\begin{proof}
We prove the result in the following steps. 

\paragraph{Step 1) Bound $\|  e^{(t)}\|_2$.}
From Lemma~\ref{lemma:multi_aget_TD}, and by applying the equation recursively, 
\begin{eqnarray*}
    e^{(t)} & = & \hat A^t e^{(0)} + \beta\alpha\gamma \sum_{l=0}^{t-1} \hat A^{t-1-l} Y + \beta\alpha \sum_{l=0}^{t-1} \hat A^{t-1-l} Z^{(l)}.
\end{eqnarray*}
From the definition of the $\ell_2$-norm, and by the triangle inequality, 
\begin{eqnarray*}
    \| e^{(t)} \|_2 
     \leq  \| \hat A \|_2^t \| e^{(0)}\|_2 + \beta\alpha\gamma \left\| \sum_{l=0}^{t-1} \hat A^{t-1-l} Y  \right\|_2+ \beta\alpha \left\| \sum_{l=0}^{t-1} \hat A^{t-1-l} Z^{(l)} \right\|_2.
\end{eqnarray*}
Next, since $\hat P_i$ is a row-stochastic matrix, its maximum eigenvalue is upper-bounded by $1$. Therefore, 
\begin{eqnarray*}
  \| \hat A  \|_2 
  & \leq & (1-\beta)+\beta \frac{1}{N}\sum_{i=1}^N \| A_i \|_2^K    \\
  & \leq & (1-\beta)+\beta \frac{1}{N}\sum_{i=1}^N [(1-\alpha)+\alpha\gamma]^K \\
  & = & (1-\beta)+\beta [(1-\alpha)+\alpha\gamma]^K.
\end{eqnarray*}
We thus have 
\begin{eqnarray*}
    \| e^{(t)} \|_2 & \leq & \rho^t \| e^{(0)}\|_2 + \beta\alpha\gamma \left\| \sum_{l=0}^{t-1} \hat A^{t-1-l} Y  \right\|_2+ \beta\alpha \left\| \sum_{l=0}^{t-1} \hat A^{t-1-l} Z^{(l)} \right\|_2,
\end{eqnarray*}
where $\rho = (1-\beta)+\beta  [(1-\alpha)+\alpha\gamma]^K$.

\paragraph{Step 2) Bound $\left\| \sum_{l=0}^{t-1} \hat A^{t-1-l} Y  \right\|_2$.}
Define $A = (1-\alpha)I+\alpha \gamma P_{\mu}$. 
Since 
\begin{eqnarray*}
Y 
& =& \frac{1}{N}\sum_{i=1}^N \sum_{l=0}^{K-1} A_i^{K-1-l} (\hat P_i - P_\mu)V\\
& = &\frac{1}{N}\sum_{i=1}^N \sum_{l=0}^{K-1} A^{K-1-l} (\hat P_i - P_\mu)V + \frac{1}{N}\sum_{i=1}^N \sum_{l=0}^{K-1} (A_i^{K-1-l}-A^{k-1-l}) (\hat P_i - P_\mu)V,
\end{eqnarray*}	
by the triangle inequality, we get
\begin{eqnarray*}
	\|  Y \|_2 
& \leq & \left\|  \frac{1}{N}\sum_{i=1}^N \sum_{l=0}^{K-1} A^{K-1-l} (\hat P_i - P_\mu)V  \right\|_2 \\
&& + \left\| \frac{1}{N}\sum_{i=1}^N \sum_{l=0}^{K-1} (A_i^{K-1-l}-A^{K-1-l}) (\hat P_i - P_\mu)V \right\|_2 \\
& \leq & \left\|  \sum_{l=0}^{K-1} A^{K-1-l} \right\|_2 \left\| \left( \frac{1}{N}\sum_{i=1}^N \hat P_i - P_\mu \right) V \right\|_2 \\
&& + \left\| \frac{1}{N}\sum_{i=1}^N \sum_{l=0}^{K-1} (A_i^{K-1-l}-A^{K-1-l}) (\hat P_i - P_\mu)V \right\|_2.
\end{eqnarray*}	
From Assumption~\ref{assum:hat_P} and by the fact that $\| V \|_2 \leq  \sqrt{ |\mathcal{S}|  } M$, we can prove that $\left\| \left( \frac{1}{N}\sum_{i=1}^N \hat P_i - P_\mu \right) V \right\|_2 \leq \left\| \frac{1}{N}\sum_{i=1}^N \hat P_i - P_\mu \right\|_2 \| V\|_2 \leq \frac{\Lambda \sqrt{|\mathcal{S}|}M}{\sqrt{N}}$, and therefore
 \begin{eqnarray*}
 	\|  Y \|_2 
 	& \leq & \left\|  \sum_{l=0}^{K-1} A^{K-1-l} \right\|_2 \frac{\Lambda \sqrt{|\mathcal{S}|}M}{ \sqrt{N} } + \left\| \frac{1}{N}\sum_{i=1}^N \sum_{l=0}^{K-1} (A_i^{K-1-l}-A^{K-1-l}) (\hat P_i - P_\mu)V \right\|_2.
 \end{eqnarray*}	
Since $\| A\| \leq 1-\alpha(1-\gamma)$ and 
\begin{eqnarray*}
	 \left\|  \sum_{l=0}^{K-1} A^{k-1-l} \right\|_2
	 & \leq &  \sum_{l=0}^{K-1} \| A \|_2^{K-1-l} \\
	 & \leq & \sum_{l=0}^{K-1} [1-\alpha(1-\gamma)]^{K-1-l} \\
	 & \leq & \sum_{l=0}^{\infty} [1-\alpha(1-\gamma)]^{l} \\
	 & = & \frac{1}{\alpha(1-\gamma)},
\end{eqnarray*}	
we obtain
\begin{eqnarray*}
\|  Y \|_2 
& \leq & \frac{\Lambda \sqrt{|\mathcal{S}|}M}{ \alpha \sqrt{N} (1-\gamma) } + \left\| \frac{1}{N}\sum_{i=1}^N \sum_{l=0}^{K-1} (A_i^{K-1-l}-A^{K-1-l}) (\hat P_i - P_\mu)V \right\|_2.
\end{eqnarray*}
Next, we can show that $A_i^l - A^l = \gamma\alpha \Psi(\hat P_i,P_\mu)$ for $\Psi(\hat P_i,P_\mu)$ is the matrix as the function of $\hat P_i,P_\mu$. Since
\begin{eqnarray*}
A_i^l - A^l
& = & [A + \alpha\gamma (\hat P_i- P_\mu)]^l -A^l \\
& = &  l A^{l-1} \cdot \alpha\gamma (\hat P_i- P_\mu) + l(l-1) A^{l-2} \frac{[\alpha\gamma (\hat P_i- P_\mu)]^2}{2!} \\
&& +  l(l-1)(l-2) A^{l-3} \frac{[\alpha\gamma (\hat P_i- P_\mu)]^3}{3!} + \ldots
\end{eqnarray*}
Therefore, 
\begin{eqnarray*}
\| A_i^l - A^l \|_2
& \leq &  l \| A \|_2^{l-1} \cdot \alpha\gamma \| \hat P_i- P_\mu\|_2 + l(l-1) \| A \|_2^{l-2} \cdot \frac{[\alpha\gamma \| \hat P_i- P_\mu\|_2]^2}{2!} \\
&& +  l(l-1)(l-2) \| A\|_2^{l-3} \frac{[\alpha\gamma \| \hat P_i- P_\mu\|_2]^3}{3!} + \ldots
\end{eqnarray*}
Since $\| A \|_2 \leq 1$ for any $\alpha\in(0,1)$, 
\begin{eqnarray*}
\| A_i^l - A^l \|_2
& \leq &  l  \cdot \alpha\gamma \| \hat P_i- P_\mu\|_2 + l^2 \frac{[\alpha\gamma \| \hat P_i- P_\mu\|_2]^2}{2!} \\
&& +  l^3  \frac{[\alpha\gamma \| \hat P_i- P_\mu\|_2]^3}{3!} + \ldots \\
& \overset{\alpha\gamma \leq 1}{\leq} & \alpha\gamma \cdot l   \| \hat P_i- P_\mu\|_2 + \alpha\gamma \cdot l^2 \frac{[ \| \hat P_i- P_\mu\|_2]^2}{2!} \\
&& +  \alpha\gamma \cdot l^3  \frac{[ \| \hat P_i- P_\mu\|_2]^3}{3!} + \ldots \\
&\leq & \alpha\gamma \left( \exp( \|l \hat P_i- P_\mu\|_2 ) -1 \right) \\
& \leq & \alpha\gamma \exp(l \| \hat P_i- P_\mu\|_2 ).
\end{eqnarray*}
Since  $\| \hat P_i^l \|_2 \leq C_P$ for some finite positive scalar $C_P$, and $\| P_\mu^l \|_2 \leq C_\mu$ for some finite positive scalar $C_\mu$, we can prove that $\| \hat P_i - P_\mu \|_2 \leq \| \hat P_i \|_2 + \| P_\mu \|_2 \leq C_P+C_\mu$, and that
\begin{eqnarray*}
\| A_i^l - A^l \|_2 \leq  \alpha\gamma \exp(l(C_P+C_\mu)).
\end{eqnarray*}
Therefore,
\begin{eqnarray*}
&& \left\| \frac{1}{N}\sum_{i=1}^N \sum_{l=0}^{K-1} (A_i^{K-1-l}-A^{K-1-l}) (\hat P_i - P_\mu)V \right\|_2\\
& &\leq  \frac{1}{N}\sum_{i=1}^N \left\|  \sum_{l=0}^{K-1} (A_i^{K-1-l}-A^{K-1-l}) \right\|_2 \left\|(\hat P_i - P_\mu)V \right\|_2 \\
& & \leq   \frac{1}{N}\sum_{i=1}^N \sum_{l=0}^{K-1}  \left\|  A_i^{K-1-l}-A^{K-1-l} \right\|_2 \left\| \hat P_i - P_\mu \right\|_2 \left\| V \right\|_2 \\
& & \leq \frac{\gamma\alpha C}{N}\sum_{i=1}^N  \left\| \hat P_i - P_\mu \right\|_2 \left\| V \right\|_2 \\
& & \leq \gamma\alpha C \Delta \sqrt{|\mathcal{S}|}M,
\end{eqnarray*}
where $C = \exp(K(C_P+C_\mu))$. Hence, 
plugging the expression of \\ $\left\| \frac{1}{N}\sum_{i=1}^N \sum_{l=0}^{K-1} (A_i^{K-1-l}-A^{K-1-l}) (\hat P_i - P_\mu)V \right\|_2$ into $\| Y\|_2$ yields
\begin{eqnarray*}
	\|  Y \|_2 
	& \leq & \frac{\Lambda \sqrt{|\mathcal{S}|}M}{ \alpha \sqrt{N} (1-\gamma) } + \gamma\alpha C \Delta \sqrt{|\mathcal{S}|}M.
\end{eqnarray*}	
Therefore, by the triangle inequality 
\begin{eqnarray*}
	\left\|  \sum_{l=0}^{t-1} \hat A^{t-1-l} Y \right\|_2
	& \leq & \sum_{l=0}^{t-1} 	\|  \hat A\|_2^{t-1-l} \| Y \|_2 \\
	& \leq & \sum_{l=0}^{t-1}  [(1-\beta) + \beta [1-\alpha(1-\gamma)]^K ]^{t-1-l} \| Y \|_2 \\
	& \leq & \| Y \|_2 \sum_{l=0}^{\infty}   [(1-\beta) + \beta [1-\alpha(1-\gamma)]^K ]^l \\
	& = & \frac{\| Y \|_2}{\beta(1-[1-\alpha(1-\gamma)]^K)} \\
	& \leq &  \frac{1}{\beta(1-[1-\alpha(1-\gamma)]^K)}\left(  \frac{\Lambda \sqrt{|\mathcal{S}|}M}{ \alpha \sqrt{N} (1-\gamma) } + \gamma\alpha C \Delta \sqrt{|\mathcal{S}|}M \right).
\end{eqnarray*}	

\paragraph{Step 3) Bound $\left\| \sum_{l=0}^{t-1} \hat A^{t-1-l} Z^{(l)} \right\|_2$ by the vector Bernstein inequality.}
Under the i.i.d. sampling regime, $\mathbb{E}[\xi_i^{(t,l)}]=0$, 
\begin{eqnarray*}
    \| \xi_i^{(t,l)} \|_2 \overset{\text{Lemma~\ref{lemma:boundedness_of_xi}}}{\leq} 2M, 
\end{eqnarray*}
and 
\begin{eqnarray*}
  \mathbb{E}   \| \xi_i^{(t,l)} \|_2 ^2   \overset{\text{Lemma~\ref{lemma:boundedness_of_xi}}}{\leq} 4M^2.
\end{eqnarray*}
From Lemma~\ref{lemma:vector_bernstein}, 
\begin{eqnarray*}
    \left\| \frac{1}{N}\sum_{i=1}^N  \xi_i^{(t,l)} \right\|_2  \leq \sqrt{\frac{32 M^2}{N}\left( \log(1/\delta_4) + 1/4\right)}
\end{eqnarray*}
with probability at least $1-\delta_4$.

Next, we derive the concentration bound of $\left\| Z^{(t)} \right\|_2$ from the concentration bound of $ \left\| \frac{1}{N}\sum_{i=1}^N  \xi_i^{(t,l)} \right\|_2 $. We can prove that $\mathbb{E}\left[ A_i^{K-1-j} \frac{1}{N}\sum_{i=1}^N\xi_i^{(t,j)} \right]=0$, 
\begin{eqnarray*}
    \left\| A_i^{K-1-j}   \frac{1}{N}\sum_{i=1}^N\xi_i^{(t,j)} \right\|_2 \leq \|  A_i \|_2^{K-1-j} \left\|   \frac{1}{N}\sum_{i=1}^N\xi_i^{(t,j)} \right\|_2 \leq \left\|   \frac{1}{N}\sum_{i=1}^N\xi_i^{(t,j)} \right\|_2, 
\end{eqnarray*}
and 
\begin{eqnarray*}
    \mathbb{E}\left\| A_i^{K-1-j}   \frac{1}{N}\sum_{i=1}^N\xi_i^{(t,j)} \right\|_2^2 \leq \|  A_i\|_2^{2(K-1-j)} \left\|   \frac{1}{N}\sum_{i=1}^N\xi_i^{(t,j)} \right\|_2^2 \leq \left\|   \frac{1}{N}\sum_{i=1}^N\xi_i^{(t,j)} \right\|_2^2. 
\end{eqnarray*}
Then, from Lemma~\ref{lemma:vector_bernstein}, 
\begin{eqnarray*}
    \| Z^{(t)}\|_2
    & = & \left\| \sum_{j=0}^{K-1}   A_i^{K-1-j}   \frac{1}{N}\sum_{i=1}^N\xi_i^{(t,j)} \right\|_2  \\
    & \leq & \sqrt{K}\sqrt{2\left\|   \frac{1}{N}\sum_{i=1}^N\xi_i^{(t,j)} \right\|_2^2(\log(1/\delta_5)+1/4)} \\
    & \leq & \sqrt{K} A(\delta_5)   \sqrt{\frac{32 M^2}{N}\left( \log(1/\delta_4) + 1/4\right)}
\end{eqnarray*}
with probability at least $1-\delta_4-\delta_5$. Here, $A(\delta)=\sqrt{2(\log(1/\delta+1/4)}$. 

Next, we derive the concentration bound of $\left\| \sum_{l=0}^{t-1} \hat A^{t-1-l} Z^{(l)} \right\|_2$ based on the concentration bound of $\| Z^{(l)}\|_2$. We can show that $\mathbb{E}[A^{t-1-l} Z^{(l)}]=0$, 
\begin{eqnarray*}
    \| \hat A^{t-1-l} Z^{(l)} \|_2 \leq \| \hat A\|_2^{t-1-l} \| Z^{(l)} \|_2 \leq \| Z^{(l)}\|_2,
\end{eqnarray*}
and 
\begin{eqnarray*}
    \mathbb{E}\| \hat A^{t-1-l} Z^{(l)} \|_2^2 \leq \| \hat A\|_2^{2(t-1-l)} \| Z^{(l)} \|_2^2 \leq \| Z^{(l)}\|_2^2.
\end{eqnarray*}
Then, from Lemma~\ref{lemma:vector_bernstein}, 
\begin{eqnarray*}
    \left\| \sum_{l=0}^{t-1} \hat A^{t-1-l} Z^{(l)} \right\|_2
    & \leq & \sqrt{t} \sqrt{2 \| Z^{(l)}\|_2^2 (\log(1/\delta_6)+1/4) } \\
    & \leq & \sqrt{t}\sqrt{K} A(\delta_5) A(\delta_6) \sqrt{\frac{32 M^2}{N}\left( \log(1/\delta_4) + 1/4\right)}
\end{eqnarray*}
with probability at least $1-\delta_4-\delta_5-\delta_6$.

\paragraph{Step 4) Derive the high-probability convergence in $\| e^{(t)} \|_2$.} By the bounds of $\left\| \sum_{l=0}^{t-1} \hat A^{t-1-l} Y  \right\|_2$ and $\left\| \sum_{l=0}^{t-1} \hat A^{t-1-l} Z^{(l)} \right\|_2$, we obtain 
\begin{eqnarray*}
    \| e^{(t)} \|_2 
     & \leq & \rho^t \| e^{(0)}\|_2 + \beta\alpha\gamma  \frac{1}{\beta(1-[1-\alpha(1-\gamma)]^K)}\left(  \frac{\Lambda \sqrt{|\mathcal{S}|}M}{ \alpha \sqrt{N} (1-\gamma) } + \gamma\alpha C \Delta \sqrt{|\mathcal{S}|}M \right) \\
    && + \beta\alpha \sqrt{t}\sqrt{K} A(\delta_5) A(\delta_6) \sqrt{\frac{32 M^2}{N}\left( \log(1/\delta_4) + 1/4\right)}
\end{eqnarray*}
with probability at least $1-\delta_4-\delta_5-\delta_6$. As each concentration bound must be valid for every $t$ and $k\in [0,K-1]$, we choose $\delta_i = \frac{\delta}{3 Kt}$ for $i=4,5,6$. We complete the proof. 
\end{proof}

\subsection{Proof of Theorem~\ref{thm:multi_markov}}
\begin{proof}
By following the proof arguments in Steps 1) and 2) of Theorem~\ref{thm:multi_iid}, we obtain 
\begin{enumerate}
    \item $\| e^{(t)} \|_2 \leq \rho^t \| e^{(0)}\|_2 + \beta\alpha\gamma \left\| \sum_{l=0}^{t-1} \hat A^{t-1-l} Y  \right\|_2+ \beta\alpha \left\| \sum_{l=0}^{t-1} \hat A^{t-1-l} Z^{(l)} \right\|_2,$ where  $\rho = (1-\beta)+\beta \frac{1}{n}\sum_{i=1}^n [(1-\alpha)+\alpha\gamma]^K$.
    \item $ \left\| \sum_{l=0}^{t-1} \hat A^{t-1-l} Y \right\|_2
   \leq \frac{1}{\beta(1-[1-\alpha(1-\gamma)]^K)}\left(  \frac{\Lambda \sqrt{|\mathcal{S}|}M}{ \alpha \sqrt{N} (1-\gamma) } + \gamma\alpha C \Delta \sqrt{|\mathcal{S}|}M \right)$.
\end{enumerate}

To complete the convergence bound in $\| e^{(t)} \|_2$, we bound $\left\| \sum_{l=0}^{t-1} \hat A^{t-1-l} Z^{(l)} \right\|_2$. 
By the triangle inequality, and by the fact that $\| A_i \| \leq 1-\alpha(1-\gamma)$ for any $\alpha \in (0,1)$, 
\begin{eqnarray*}
\| Z^{(t)} \|_2 
& \leq &  \frac{1}{N}\sum_{i=1}^N\sum_{j=0}^{K-1} \left\| A_i \right\|_2^{K-1-j} \left\| \xi_i^{(t,j)} \right\|_2 \\
& \leq &  \frac{1}{N}\sum_{i=1}^N\sum_{j=0}^{K-1} (1-\alpha(1-\gamma))^{K-1-j}\left\| \xi_i^{(t,j)} \right\|_2 \\
& \leq &  \frac{1}{N}\sum_{i=1}^N\sum_{j=0}^{\infty} (1-\alpha(1-\gamma))^{K-1-j}\left\| \xi_i^{(t,j)} \right\|_2 \\
& \leq & \frac{1}{\alpha(1-\gamma)} \frac{1}{N}\sum_{i=1}^N \max_{j \geq 0}\left\| \xi_i^{(t,j)} \right\|_2.
\end{eqnarray*}
Therefore, by the fact that $\| \hat A\|_2 \leq 1$ for any $\beta \in (0,1/[1-[1-\alpha(1-\gamma)]^K])$, 
\begin{eqnarray*}
\left\| \sum_{l=0}^{t-1} \hat A^{t-1-l} Z^{(l)} \right\|_2
& \leq & \sum_{l=0}^{t-1} \|  \hat A \|_2^{t-1-l} \| Z^{(l)} \|_2 \\	
& \leq & \sum_{l=0}^{t-1} \| Z^{(l)} \|_2 \\ 
& \leq & \frac{1}{\alpha(1-\gamma)}\frac{1}{N}\sum_{i=1}^N \sum_{l=0}^{t-1}  \max_{j \geq 0} \left\| \xi_i^{(l,j)} \right\|_2.
\end{eqnarray*}	
Next, by the definition of the mixing time (Definition~\ref{def:mixing_time_federated}), 
\begin{eqnarray*}
	\left\| \sum_{l=0}^{t-1} \hat A^{t-1-l} Z^{(l)} \right\|_2
	& \leq & \frac{1}{\alpha(1-\gamma)}\frac{1}{N}\sum_{i=1}^N \sum_{l=0}^{\tau_{\epsilon}-1}  \max_{j \geq 0} \left\| \xi_i^{(l,j)} \right\|_2 \\
	&& +  \frac{1}{\alpha(1-\gamma)}\frac{1}{N}\sum_{i=1}^N \sum_{l=\tau_{\epsilon}}^{t-1}  \max_{j \geq 0} \left\| \xi_i^{(l,j)} \right\|_2 \\
	& \leq &  \frac{1}{\alpha(1-\gamma)}\frac{1}{N}\sum_{i=1}^N \sum_{l=0}^{\tau_{\epsilon}-1} 2M +  \frac{1}{\alpha(1-\gamma)}\frac{1}{N}\sum_{i=1}^N \sum_{l=0}^{t-1}  \epsilon \\
	& \leq & \frac{1}{\alpha(1-\gamma)}(2M \tau_\epsilon + t \epsilon).
\end{eqnarray*}	    
Thus, we obtain 
\begin{eqnarray*}
	\| e^{(t)} \|_2 
	& \leq & \rho^t \| e^{(0)}\|_2 + \beta\alpha\gamma  \frac{1}{\beta(1-[1-\alpha(1-\gamma)]^K)}\left(  \frac{\Lambda \sqrt{|\mathcal{S}|}M}{ \alpha \sqrt{N} (1-\gamma) } + \gamma\alpha C \Delta \sqrt{|\mathcal{S}|}M \right)  \\
	&& + \beta\alpha\frac{1}{\alpha(1-\gamma)}(2M \tau_\epsilon + t \epsilon).
\end{eqnarray*}
Finally, by choosing $\epsilon=\beta$, we complete the proof. 
\end{proof}

\section{Additional Experiments} \label{sec:AdditionalExperiments}
\begin{figure*}[!htb] 
\centering
\subfloat[Number of agents $N=1$\label{figAppendix:delta_iid_N1}]{
  \includegraphics[width=0.49\textwidth]{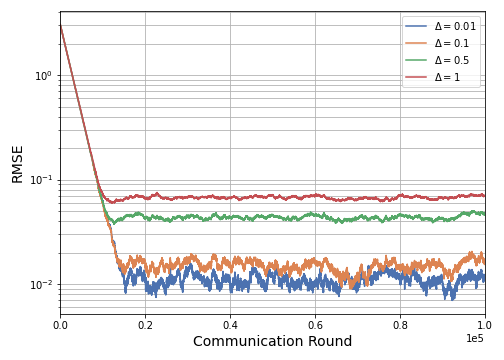}}
\subfloat[Number of agents $N=10$\label{figAppendix:delta_iid_N10}]{
  \includegraphics[width=0.49\textwidth]{plots/fedtd_n_10_N_10_gamma_0.8_alpha_0.01_beta_0.4_T_100000_K_5_s_iid_iidopt_stationary_R_uniform.png}}
  \hfil
\subfloat[Number of agents $N=20$\label{figAppendix:delta_iid_N20}]{
  \includegraphics[width=0.49\textwidth]{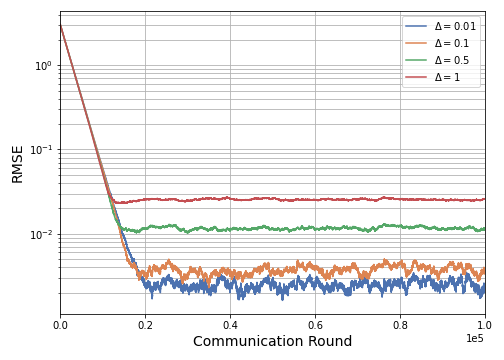}}
\subfloat[Number of agents $N=100$\label{figAppendix:delta_iid_N100}]{
  \includegraphics[width=0.49\textwidth]{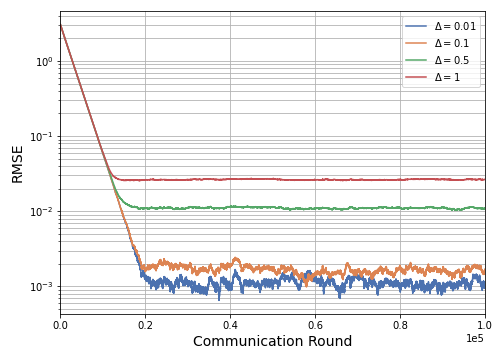}}
\caption{Impact of the model mismatch $\Delta$: the RMSE of the $V$-estimates for \textsf{FedTD(0)} with the i.i.d sampling. Here, the number of local steps $K=5$, the learning rate $\alpha=0.01$, and the federated parameter $\beta=0.4$.}
\label{figAppendix:delta_iid}
\end{figure*}
\begin{figure*}[!htb] 
\centering
\subfloat[Number of agents $N=1$\label{figAppendix:delta_markov_N1}]{
  \includegraphics[width=0.49\textwidth]{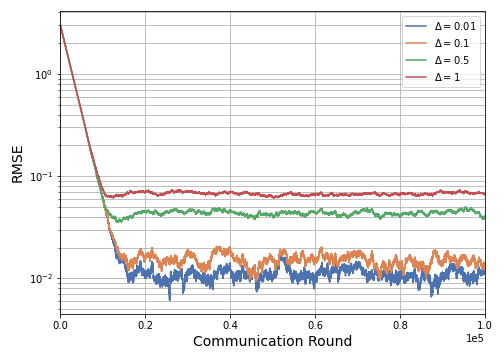}}
\subfloat[Number of agents $N=10$\label{figAppendix:delta_markov_N10}]{
  \includegraphics[width=0.49\textwidth]{plots/fedtd_n_10_N_10_gamma_0.8_alpha_0.01_beta_0.4_T_100000_K_5_s_markov_iidopt_uniform_R_uniform.png}}
  \hfil
\subfloat[Number of agents $N=20$\label{figAppendix:delta_markov_N20}]{
  \includegraphics[width=0.49\textwidth]
{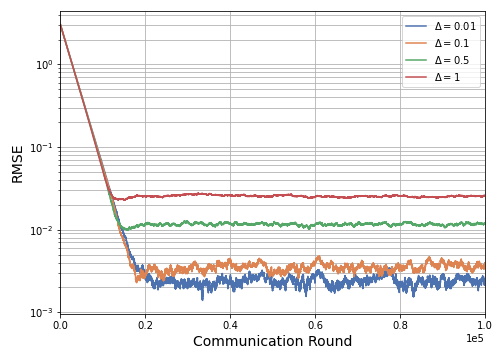}}
\subfloat[Number of agents $N=100$\label{figAppendix:delta_markov_N100}]{
  \includegraphics[width=0.49\textwidth]{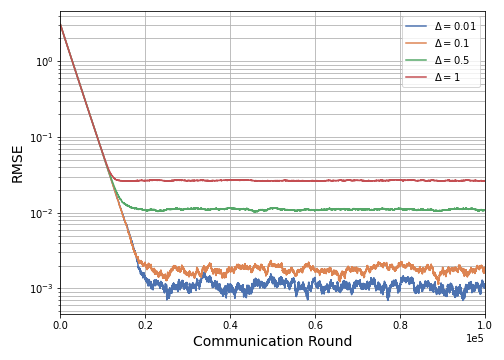}}
\caption{Impact of the model mismatch $\Delta$: the RMSE of the $V$-estimates for \textsf{FedTD(0)} with Markovian sampling. Here, the number of local steps $K=5$, the learning rate $\alpha=0.01$, and the federated parameter $\beta=0.4$.}
\label{figAppendix:delta_markov}
\end{figure*}
\begin{figure*}[!htb] 
\centering
\subfloat[The model mismatch $\Delta=0.01$\label{figAppendix:N_markov_delta0.01}]{
  \includegraphics[width=0.49\textwidth]{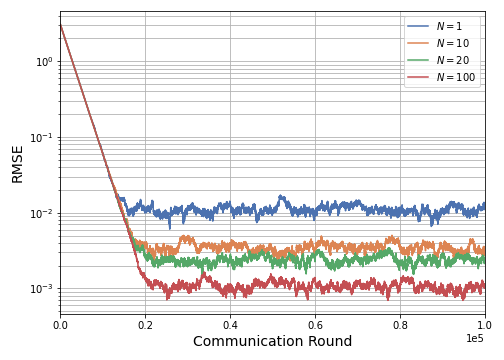}}
\subfloat[The model mismatch $\Delta=0.1$\label{figAppendix:N_markov_delta0.1}]{
  \includegraphics[width=0.49\textwidth]{plots/fedtd_n_10_Delta_0.1_gamma_0.8_alpha_0.01_beta_0.4_T_100000_K_5_s_markov_iidopt_uniform_R_uniform.png}}
  \hfil
\subfloat[The model mismatch $\Delta=0.5$\label{figAppendix:N_markov_delta0.5}]{
  \includegraphics[width=0.49\textwidth]
{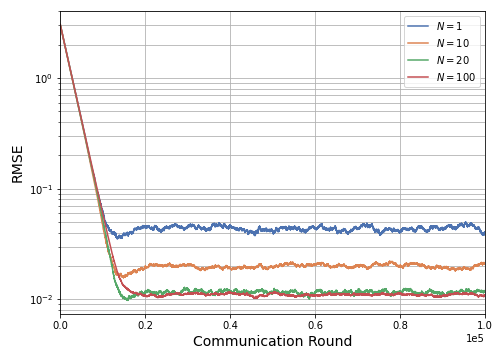}}
\subfloat[The model mismatch $\Delta=1$\label{figAppendix:N_markov_delta1.0}]{
  \includegraphics[width=0.49\textwidth]{plots/fedtd_n_10_Delta_1.0_gamma_0.8_alpha_0.01_beta_0.4_T_100000_K_5_s_markov_iidopt_uniform_R_uniform.png}}
\caption{Impact of the number of agents $N$: the RMSE of the $V$-estimates for \textsf{FedTD(0)} with Markovian sampling. Here, the number of local steps $K=5$, the learning rate $\alpha=0.01$, and the federated parameter $\beta=0.4$.}
\label{figAppendix:N_markov}
\end{figure*}
In this section, we provide additional experimental results that further support our main findings on the effect of model mismatch, the number of agents, and the choice of sampling regime on the performance of \textsf{FedTD(0)}.

These additional results further validate our theoretical analysis and highlight the robustness of \textsf{FedTD(0)} in handling transition mismatches across both i.i.d.\ and Markovian sampling settings.

\subsection{Effect of Model Mismatch under I.I.D. Sampling} \label{sec:AdditionalExperiments_IID}
Figure~\ref{figAppendix:delta_iid} examines the impact of model mismatch $\Delta$ on the RMSE of the $V$-estimates in the i.i.d.\ setting for different numbers of agents $N$. The results show that while larger $\Delta$ leads to increased residual error, increasing $N$ consistently reduces this error, aligning with our theoretical predictions. Notably, even with a severe model mismatch, federated updates help mitigate the bias introduced by environmental heterogeneity.

\subsection{Effect of Model Mismatch under Markovian Sampling} \label{sec:AdditionalExperiments_Markov}
Similar to the i.i.d.\ case, Figure~\ref{figAppendix:delta_markov} illustrates the effect of model mismatch under Markovian sampling. The trends remain consistent, confirming that increasing $\Delta$ enlarges the bias but can be counteracted by increasing $N$. These results reinforce our findings in Theorems \ref{thm:multi_iid} and \ref{thm:multi_markov}, which predict that federated averaging helps stabilize learning under both sampling regimes.

\subsection{Effect of Model Mismatch Across Different $\Delta$} \label{sec:AdditionalExperiments_Delta}
Figure~\ref{figAppendix:N_markov} provides a detailed analysis of how different values of $\Delta$ affect \textsf{FedTD(0)} under Markovian sampling. As expected, the RMSE increases with $\Delta$, but the impact diminishes as $N$ grows. This is in direct agreement with Theorems \ref{thm:multi_iid} and \ref{thm:multi_markov}, which state that larger $N$ decreases the error and mitigates the mismatch effect. The results confirm that even in highly heterogeneous environments, increasing $N$ significantly improves convergence accuracy.

\end{document}